\documentclass[5p]{elsarticle}
\usepackage{lineno,hyperref}
\modulolinenumbers[5]

\usepackage[ruled,vlined]{algorithm2e}
\usepackage{makecell}
\usepackage{multirow}
\usepackage{amsmath}
\usepackage{amsfonts}
\usepackage{enumitem}
\usepackage{pifont}
\usepackage{arydshln}
\usepackage{xcolor}
\graphicspath{{figures/}}

\journal{ISPRS Journal of Photogrammetry and Remote Sensing, }
\date{Sept 30, 2021}

\biboptions{authoryear}

\begin{document}

\begin{frontmatter}

\title{LiDAR Point--to--point Correspondences \\ for Rigorous Registration of Kinematic Scanning \\ in Dynamic Networks}

\author[topo]{Aurélien Brun}
\author[unige]{Davide A. Cucci\corref{corauthor}}
\ead{davide.cucci@unige.ch}
\author[topo]{Jan Skaloud}
\ead{jan.skaloud@epfl.ch}

\cortext[corauthor]{Corresponding author}
\address[topo]{EPFL ENAC TOPO, Bâtiment GC - Station 18, 1015 Lausanne, Switzerland} 
\address[unige]{Geneva School of Economics and Management, University of Geneva, Bd du Pont-d'Arve 40, 1205, Gen\`eve, Switzerland}

\begin{abstract}
With the objective of improving the registration of LiDAR point clouds produced by kinematic scanning systems, we propose a novel trajectory adjustment procedure that leverages on the automated extraction of selected reliable 3D point--to--point correspondences between overlapping point clouds and their joint integration (adjustment) together with all raw inertial and GNSS observations. This is performed in a tightly coupled fashion using a Dynamic Network approach that results in an optimally compensated trajectory through modeling of errors at the sensor, rather than the trajectory, level. The 3D correspondences are formulated as static conditions within this network
and the registered point cloud is generated with higher accuracy utilizing the corrected trajectory and possibly other parameters determined within the adjustment. We first describe the method for selecting correspondences and how they are inserted into the Dynamic Network as new observation models. We then describe the experiments conducted to evaluate the performance of the proposed framework in practical airborne laser scanning scenarios with low-cost MEMS inertial sensors. In the conducted experiments, the method proposed to establish 3D correspondences is effective in determining point--to--point matches across a wide range of geometries such as trees, buildings and cars. Our results demonstrate that the method improves the point cloud registration accuracy, that is otherwise strongly affected by errors in the determined platform attitude or position (in nominal and emulated GNSS outage conditions), and possibly determine unknown boresight angles using only a fraction of the total number of 3D correspondences that are established.
\end{abstract}


\end{frontmatter}

\section{Introduction}
\subsection{Challenges in kinematic laser scanning}
Thanks to active sensing that increases the discrimination of fine elements, an ability to “peer through” vegetation and directly reconstruct three-dimensional surfaces, kinematic laser scanning remains a key technology to provide foundational data for the creation of digital twins of infrastructure (both outdoor and indoor)cities as well as high-resolution terrain models. Together with its independence on external illumination as an active sensor, these properties make mobile laser scanning (MLS)indispensable for rapid digitalization of fine features in urban infrastructure and airborne laser scanning (ALS) the essential tool for deriving digital elevation models, precise quantification of above ground biomass, canopy structure, snow accumulation, detection of stream networks, high-resolution bathymetry, etc. However, this modern technology is not without its disadvantages. Under certain circumstances, due to its need for direct orientation, kinematic laser scanning is challenging in dense urban environments, indoors and in areas where GNSS signal occlusions occur in general; as well as on small UAVs or high-altitude carriers where the influence of attitude noise on point cloud registration surpasses that of ranging accuracy. 
\begin{figure}[h!] 
\centering 
\includegraphics[width = 8cm]{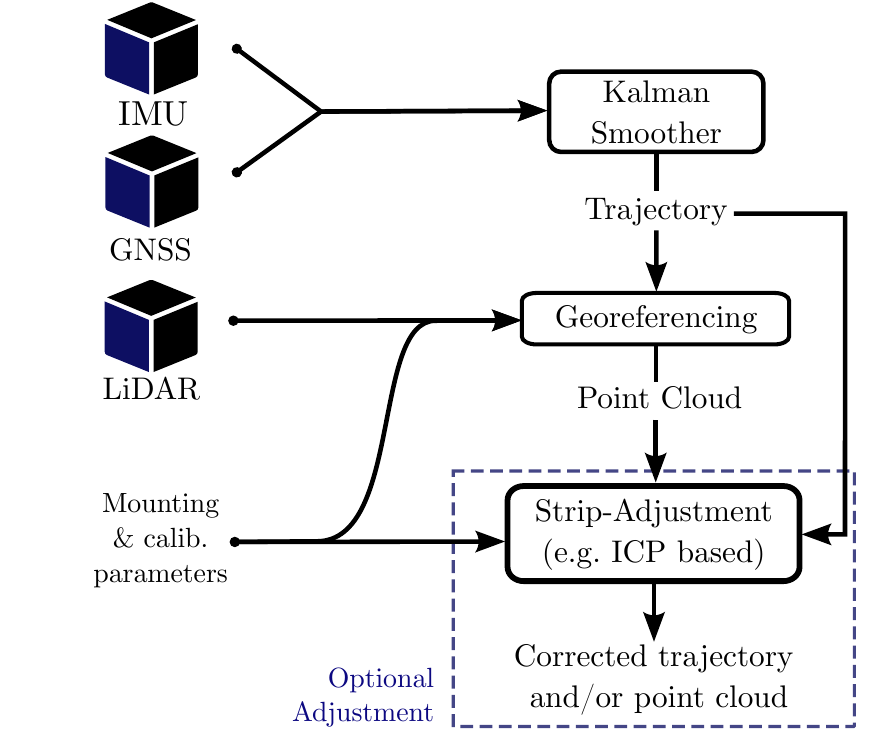} 
\caption{\label{fig:std_pipeline} Overview of the standard georeferencing based on Kalman smoother and strip adjustment.} 
\end{figure}
A standard processing scheme of kinematic laser scanning is depicted in Fig.~\ref{fig:std_pipeline}. As discussed in detail by \cite{Glennie2007a}, the accuracy of individual points is, apart from ranging quality, affected by the forward propagation of errors starting from the trajectory determination by navigation sensors (outdoors mainly INS/GNSS), the incertitude in mounting (lever-arm, boresight) and the parameters related to scanner interior orientation. These, together with sampling density, influence the geometrical quality of derived elevation models \citep{SkaloudSchaer2012} or other objects of interest. 
The prevailing source of error depends on the application and sensor quality. Generally, indoor and MLS applications are more influenced by position error due to GNSS signal occlusion; while ALS is more affected by the projection of attitude errors in the case of ``long'' ranging distances. 

\subsection{Related work}

When the accumulated errors in kinematic laser systems are higher than the ranging noise, the discrepancies between scans from different perspectives (directions, altitude, etc.) become noticeable and thus the need to mitigate them arises. This holds for conventional mapping as well as for specially designed ``in-motion'' calibration scenarios. In the latter case, the aspects of trajectory quality are usually controlled through planning, so the subsequent adjustment can be formulated in a rigorous way either with ``good enough'' trajectories as in e.g., \cite{Kager2004, Filin2003, Skaloud2006b, Friess2006, Kerstling2012}, or with minimum assumptions on the trajectory deficiency as in e.g.,  \cite{Filin2004, Hebel2012, GliraRigICP2015}. In the former situation, trajectory corrections are often modelled as angular and/or position offsets that are considered time-invariant either per block or per strip/line.
While these assumptions are possibly reasonable for situations in which laser scanners are mounted on stabilized platforms with high quality IMUs, they do not hold in general. For instance, in MLS the shape of the trajectory due to errors in position may change quickly and non-linearly in GNSS-denied environments \citep{SchaerEurocow2016}. For non-stabilized lasers on platforms such as helicopters or Unmanned Aerial Vehicles~(UAVs) that do not necessarily fly in regular grid pattern (e.g., with a constant velocity and minimal angular change during flight lines), the projection of orientation-induced errors varies substantially within a strip. Furthermore, in lightweight UAVs employing industrial-grade IMUs, the error signatures of point clouds created by attitude-projected errors become ``too wavy'' to be captured by such simple modelling \citep{ValletPuck2020}.

In short, in many cases the deficiencies in the estimated trajectory cause time-dependent, non-linear, and possibly fast-changing deformations of point clouds during data collection. To mitigate such errors one can either attempt to model their \textit{effect} on the trajectory, as proposed for instance by \cite{GliraAdjTimeErr2016}, or directly their \textit{cause} as put forward by this contribution. We argue that while the former approach may work in certain scenarios, the latter is rigorous and thus can be applied more generally. Furthermore, and as will be demonstrated practically, the modelling and estimation of the \textit{cause} requires a relatively small number of 3D tie-features to correct errors in the trajectory (and indirectly in the point cloud) that otherwise have complex appearances in the mapping frame. 

A good selection of tie-features is an important prerequisite for the proposed method to be applied during general-operation as well as during in-motion calibration.
The parameter recovery of non-calibrated kinematic scanning systems usually benefits from some a-priory knowledge on the existing physical form of scanned patches (e.g., planarity,etc.) as in \cite{Kager2004, Filin2003, Friess2006, Glennie2010}. When coupled with a strong observation geometry, such conditioning tolerates poor initial registration quality, guarantees parameter observability \citep{Skaloud2006b} and allows automation \citep{LIBOR}. Nevertheless, this approach is better suited for calibration than for general mapping where the a-priory assumption on the presence of certain surface geometries cannot be made. Hence, the conditioning is usually formulated as some surface to surface intrinsic, point to surface patch \citep{Kerstling2012} or point--to--point. The last option usually stems from some evolved variants of the Iterative Closest (or Corresponding) Point (ICP) algorithm, independently proposed by \cite{BeslICP1992} and \cite{ChenICP1991} for registration between terrestrial scanning stations. As point--to--point minimization is practically simpler to deal with, it is also widely applied in ALS~\citep{GliraPFG2015}. Nevertheless, this approach exhibits only a local linear convergence \citep{Pottmann2006} and therefore requires a good approximation of the initial registration; a case that may not hold in UAV based ALS or MLS due to previously mentioned reasons. In the proposed methodology we assume that only some point--to--point correspondences can be found within the overlapping regions of measured point clouds based on the analysis of their surroundings. In this respect, our goal is not to propose a new algorithm to establish such correspondences, but rather to exploit and adapt some recent advances in this domain when utilizing it within the Dynamic Networks adjustment approach. Although we previously studied the potential benefits of employing a common modeling-estimation approach in kinematic laser scanning for trajectory estimation within a simulated scenario \citep{Rouzaud2011}, 
the proposed adjustment approach involves basic steps introduced within our former work in photogrammetry \citep{cucci_rawbundle_2017} while expanding it to make use of 3D point--to--point correspondences (derived from an approximated point cloud) instead of 2D tie-points derived from images. 

\subsection{Proposed approach}

\begin{figure}[h!] 
\centering 
\includegraphics[width=8cm]{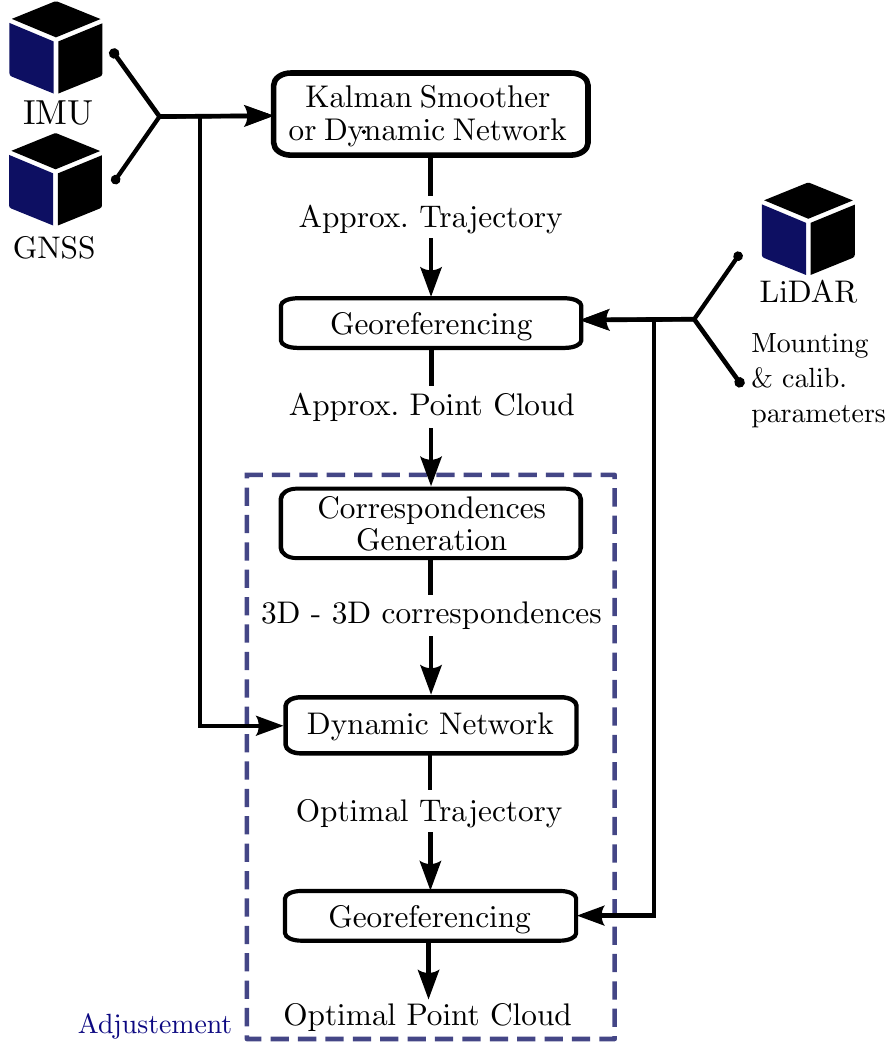}
\caption{\label{fig:pipeline} Overview of the proposed georeferencing procedure.}
\end{figure}

We adapt the final stage of the formerly described kinematic laser scanning registration methodology  (Fig.~\ref{fig:std_pipeline}). There we propose to utilize raw inertial and GNSS observations together with a percentage of the laser measurements within the overlapping point cloud regions in a common adjustment step as schematically shown in Fig~\ref{fig:pipeline}. Thanks to the rigorous modeling of observations and error sources on the sensor and system level that this approach yields, an optimal registration of the whole trajectory is achieved. The obtained optimal trajectory (and possibly other system parameters) are then applied in the final registration (georeferencing) of the measured point clouds.

The remainder of the paper is organised as follows. In Section $2$ we detail the correspondence generation procedure. In Section $3$ we present how the generated point--to--point correspondences are employed within Dynamic Networks, along with other raw sensor measurements, to compute an optimal trajectory.
In Section $4$ we describe the design and implementation of experiments that were established to assess the merits of the proposed method with respect to high-quality references in position, attitude and final registered point clouds. In Section $5$ we discuss the results of such experiments. We first consider a nominal scenario where we correct the registration errors  (mainly) caused by the noise of the MEMS-IMU attitude over distances that are expected to be achieved by modern UAV based laser scanners \citep{minivux}. We then study the possibility of recovering some calibration parameters, such as an imperfect LiDAR boresight, within the adjustment process. Finally, we look at the capacity of the method to correct errors in the point cloud registration for a trajectory where GNSS outages occur.
\section{LiDAR correspondences}\label{sec:correspondences}
\begin{figure*}[h!] 
\centering 
\includegraphics[width = 14 cm]{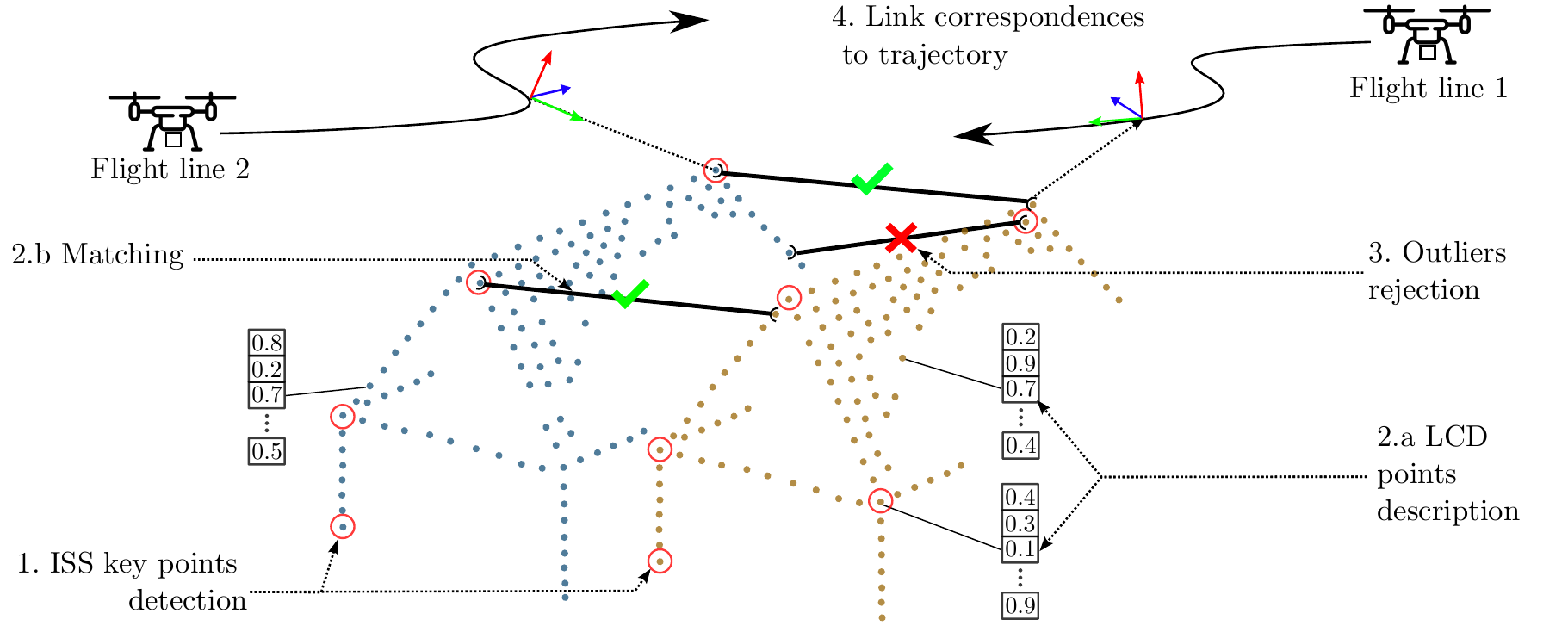}
\caption{\label{fig:matching}Overview of steps to connect trajectories via LiDAR pulse observations through the \textit{key point--to--point} detection and matching principle. Two overlapping and misaligned LiDAR point clouds in blue and orange are related to a portion of two overlapping flight lines $1$ and $2$.}
\end{figure*}

\subsection{Preprocessing}

The first step is to extract the overlapping region between two point clouds. This region is then fragmented into rectangular tiles of tractable size that are stored in pairs. The steps that follow within the correspondence estimation process are performed independently on each tile pair.

\subsection{Key point detection}
3D key point detection aims at selecting salient points from a point cloud (edges, corners) where their distinctiveness with respect to their neighborhood is expected to ease matching with corresponding points in other point clouds. We do not aim to design a detector, rather we have evaluated several state-of-the-art detectors on close-range ALS data (ranges $<$ 300 m) and applied the most suitable one. We have selected the ISS (Intrinsic Shape Signature) algorithm \citep{ISS} to detect key points because of its robustness and performance compared to the other algorithms tested, as mentioned in \cite{detector_eval}. ISS was originally designed to be a descriptor but its first 3 steps can be used independently to detect key points. We use its open source implementation from the Point Cloud Library \citep{PCL}, which enables the detection of key points in a few seconds at the scale of a fragment. In this implementation, only the first four \\

ISS employs the following steps to detect key points: 
\begin{enumerate}[noitemsep]
    \item For every point, its spherical neighborhood is extracted and a weight inversely proportional to its number of neighbors is computed.
    \item For every point, a weighted covariance matrix of its neighborhood is computed and its three eigen values are extracted.
    \item Points that show high ratio in the successive eigen values of their covariance matrix are selected as key point candidates.
\end{enumerate}
The complete formulation of the algorithm can be found in the original publication as well as in the PCL source code \citep{PCL}.
The detection is performed separately on the two point clouds of each tile pair, returning for each of these pairs two independent sets of key points.

\subsection{Point description and matching}
The aim of point description is to compute a lower dimensionality representation of a point and its surroundings or, in other words, to "encode its (geometric) neighborhood" into a feature vector so that similar structures have similar feature vectors. It is then possible to match points by comparing their feature vectors using a distance metric, e.g., the L2 norm of the difference of the two descriptors.

Point cloud description is a challenging task due to the sparse nature of point clouds originating from kinematic laser scanning (i.e., compared to images) and to their high variability depending on the scanned surface and the technology and vehicle used. Substantial effort has been devoted to the design efficient point cloud descriptors, especially since the advent of deep-learning, which has brought about rapid improvement in this task \citep{learning_based_overview, descriptor_overview}. These descriptors are usually designed for TLS or MLS point clouds and their generalization potential to ALS point clouds is uncertain. Therefore we tested multiple 3D descriptors, namely FPFH \citep{FPFH}, SHOT \citep{SHOT}, USC \citep{USC}, SpinNet \citep{spinnet}, FCGF \citep{FCGF} and LCD \citep{lcd}. While not being fully detailed here, we provide some evaluation of their suitability with respect to our ALS experimental data in Appendix 1. \\
In the case of close-range ALS over a mixed natural-built environment, LCD showed the best descriptiveness and robustness thus it was selected for use in our final implementation. LCD is a siamese neural network originally designed for 2D to 3D matching between images and point clouds but the neural network responsible for description of 3D points can be used independently. LCD is able to establish LiDAR correspondences on a diverse range of surfaces (e.g., man made constructions and vegetation). We used the open source pre-trained version of the network (trained on indoor point clouds). LCD works as follows: given a key-point, its spherical neighborhood is extracted and for each neighbor, its relative coordinates are computed by taking the key-point as the origin of the sphere. The neighborhood is then fed to the neural network, which outputs a feature vector for the point.\\
This representation of the neighborhood ensures translation invariance of the descriptor, meaning that the output feature vector is the same if the point cloud is translated by any vector. This property allows for point clouds to be matched regardless of their misalignment, by solely relying on their intrinsic geometry. Nevertheless, LCD is not a rotation invariant descriptor, meaning that the matching will fail if one of the point cloud is strongly rotated with respect to the other. Even with trajectories of lower quality, the global orientation of an associated point cloud is usually sufficient -- for the purpose of descriptors -- since attitude errors of the trajectory are projected mostly in planimetry. \\
Once feature vectors are computed for every point of a tile pair, the following matching strategy is performed. For every key-point in a given reference tile, we find its nearest neighbor in the query tile, using the Euclidean distance as a feature vector similarity metric, generating what we call a set of point--to--point correspondences. Note that, contrary to common matching procedures where correspondences are established comparing key-points to key-points, we compare key-points with all other points from the second cloud. This is because when matching key-points to key-points, a correspondence can be generated from a distinctive feature (e.g., the edge of a building) only if it is detected precisely as a key-point in both point clouds. In other words, the detector must be able to retrieve the exact same key points between the two point clouds of a scene. This characteristic is called detector "repeatability" \citep{detector_eval} and is hard to obtain in practice. Repeatability is especially challenging to achieve with ALS data because of the difference in laser sampling density and the variable viewpoints at which ALS flight lines can be flown. The matching proposed here mitigates the dependency on detector repeatability. This comes at the price of obtaining a larger number of candidates and subsequently increasing the risk of outliers, which together motivate the design of an efficient outlier filtration approach.

\subsection{Outlier rejection}
The situation with detected candidates between two overlapping point clouds is somewhat similar to image matching, where even with powerful descriptors and matching procedures, one still ends up with sets of correspondences contaminated by outliers. 
We utilize RANSAC \citep{RANSAC} as the base of our outlier filtration algorithm for its efficiency in processing moderately to highly contaminated datasets. The assumptions applied in the filtering process are as follows:\\
In ALS, the error in vehicle position and attitude will vary in time during the flight, causing georeferencing errors to vary in the point cloud along the flight lines. This implies that the misalignment between two point clouds is not rigid and changes along the overlapping region. The amplitude of these changes is nevertheless limited to the scale of a small tile (e.g., 50 by 50 meters) since it is scanned in a few seconds by the aerial platform. During such a short time span, the navigation errors i) are mainly caused by IMU noise, ii) are strongly correlated and iii) have a similar impact on the entire tile. This observation allows us to establish the following statement:
Given a set of contaminated correspondences, there exists a 6 DoF transformation that approximately aligns the correct correspondences, but which does not align the incorrect correspondences (outliers). The inliers are only approximately aligned because of the slight changes in misalignment at the scale of a tile (e.g., non-rigid misalignment) and because of the discrete nature of the point cloud. This makes it necessary to use a qualitative criteria to separate inliers and outliers. Formally, given the correct 6 DoF transformation (\textbf{T} and \textbf{R}) and two points of a correspondence ($p_a$ and $p_b$), this correspondence is classified as an inlier if the following condition is true:\\
\begin{equation}
    ||\textbf{R}p_a+\textbf{T}-p_b||< \tau 
    \label{eq:ransacth}
\end{equation} 
In other words, a correspondence is classified as an inlier if the misalignment between the two points after applying the transformation is below a certain threshold $\tau$. This approach allows the non rigid nature of the misalignment to be accounted for (something that is not possible using ICP). This is the model tested in the implemented RANSAC-based algorithm, that is defined as follows: 

\begin{itemize}[noitemsep]
    \item \textbf{C} the complete set of \textbf{K} correspondences. 
    \item $\mathbf{p_a^k}$ and $\mathbf{p_b^k}$ the first and second point of the $k^{th}$ correspondence.
    \item \textbf{R} and \textbf{T} the rotation matrix and translation vector defining the 6 DoF transformation.
    \item $\mathbf{\tau}$ the tolerance threshold to classify correspondences as inliers or outliers given an estimated 6 DoF transformation parameter.
    \item \textbf{s} the number of correspondences to select to form a subset.
    \item \textbf{N$_{it}$} the number of iterations to perform, defined following the probabilistic approach from the original publication \citep{RANSAC}.
\end{itemize}

\begin{algorithm}[h!]
\SetAlgoLined
\SetKwInOut{Parameter}{Parameter}
\KwData{Set of correspondences $\mathbf{C}$}
\Parameter{$\mathbf{\tau},N_{it},\mathbf{s}$}
\While{t $<$ N$_{it}$}{
    Randomly select $\mathbf{C_{sub}} \subset \mathbf{C}$ of $\mathbf{s}$ elements\\
    Estimate $\hat{\mathbf{R}}, \hat{\mathbf{T}}$:   \begin{center}
        
     $\hat{\mathbf{R}}, \hat{\mathbf{T}} = \underset{R,T}{argmin}\sum_{\mathbf{i}}^{\mathbf{s}}||\mathbf{R}p_a^i+\mathbf{T}-p_b^i||  $\end{center}
    \For{ $\mathbf{c} \in \mathbf{C} \setminus \mathbf{C_{sub}}$}{
        \uIf{$||\hat{\mathbf{R}}p_a^c+\hat{\mathbf T}-p_b^c||< \tau$}{c is an inlier}
        \uElse{c is an outlier}
        
    }
    Inlier ratio$_t = \frac{\# Inliers}{\# Correspondences}$\\
    t++
 }
Results: $\hat{\mathbf{R}}, \hat{\mathbf{T}}$ and inlier classification for the subset with most inliers
\caption{RANSAC}
\end{algorithm}

The output of such algorithm includes the set of correspondences that satisfy Equation~\ref{eq:ransacth} for $\hat{\mathbf{R}}$ and $\hat{\mathbf{T}}$. The algorithm is applied to all tiles where overlap exists between two point clouds. The complete set of inlier correspondences is used in the subsequent trajectory adjustment within Dynamic Networks, as will be discussed in the next section.


\section{Dynamic Network with LiDAR constraints}\label{sec:DN+C}

Dynamic Networks~(DNs) are an extension of conventional geodetic networks and were first introduced in~\citep{blazquez2004, SkaloudGAL2015}. In DNs, a hyper-graph is constructed where the nodes are the unknown body frame position and orientation at discrete time instants, $\Gamma^n_{b, t}$, calibration parameters (e.g., the LiDAR boresight, $R^b_L$) and inertial sensor biases. Hyper-edges (edges connecting multiple nodes) encode constraints between the connected unknowns given by sensor measurements. Such hyper-graph is a graphical model, known as factor-graph~\citep{loeliger2007factor}, of the joint probability distribution of the observed sensor measurements given the unknowns. An optimal estimation of the unknowns can thus be obtained by determining the values that maximize the joint likelihood of the sensor measurements.  This corresponds to modern formulations of the Simultaneous Localization and Mapping~(SLAM) problem in robotics, see for example~\citep{cioffi2020tightly}.

Typically, we assume that all the random effects are zero mean and Gaussian distributed. In this setting, each hyper-edge computes the difference between the predicted sensor measurement, given the connected unknowns (via a measurement model), and the observed one. The max-likelihood estimator for the unknowns is then obtained minimizing all such differences (residuals) by solving a non-linear (optionally robust) weighted least-squares optimization problem, for example employing a Gauss-Newton minimization algorithm. This work is based on the formulation presented in detail in~\citep{cucci_rawbundle_2017}, and is extended to include LiDAR correspondences, as will be discussed in the following, as well as the more sophisticated inertial sensor modeling introduced in~\citep{cucci2019onraw}. In the latter, a gravity model, the Earth rotation and IMU pre-integration are considered in inertial sensor models to remove limitations on the size of the mapped area.

\subsection{LiDAR correspondences error model}

In order to consider LiDAR correspondences in DNs, it is necessary to specify the measurement model for one such correspondence.
Let us assume that the algorithm presented in Section~\ref{sec:correspondences} has established that a certain point $p_1$, found in one point cloud, corresponds to point $p_2$, found in an overlapping one. These points can be traced back to the original LiDAR measurements, corresponding to two vectors $v^L_1$ and $v^L_2$, expressed in LiDAR frame $L$, acquired at $t_1$ and $t_2$. Since $p_1$ and $p_2$ are corresponding points, $v^L_1$ and $v^L_2$ should match once expressed in the navigation frame $n$. This can be formally expressed as follows:
\begin{equation}
    \Gamma^n_b(t_1) R^b_L v^L_1 - \Gamma^n_b(t_2) R^b_L v^L_2 = 0 + \xi,\;\; \xi \sim \mathcal{N}(0, \sigma\text{ I}_{3\times3}).
    \label{eq:lc1}
\end{equation}
where $\Gamma^n_b(t) \in \text{SE}(3)$ is the \emph{continuous} time body frame position and orientation (pose) at time $t$ and $R^b_L$ is the LiDAR boresight. Here, the product operator corresponds to the usual composition of rigid transformations. 
$\xi$ is a zero-mean Gaussian noise whose variance has to be adjusted depending on the point cloud density and the assumed statistics of the error in the generated correspondences.

In DNs, body frame poses are kept at discrete time instants, i.e., $\Gamma^n_{b,t}$, synchronous with the IMU. Since the rate of typical LiDAR sensors is substantially higher, $t_1$ and $t_2$ are arbitrary and $\Gamma^n_b(\cdot)$, as it appears in Equation~\ref{eq:lc1}, is not available in general. Thus, we replace $\Gamma^n_b(t_1)$ and $\Gamma^n_b(t_2)$ with two new poses, $\tilde \Gamma^n_{b,t_1}$ and  $\tilde \Gamma^n_{b,t_2}$, which are constrained to lie on the geodetic in $\text{SE}(3)$ between the nearest poses available in the DN. This approach is described in detail in~\citep[Section 4.2]{cucci_rawbundle_2017}, where it was employed to handle image observations at arbitrary timestamps, and in~\citep{ceriani2015pose} in the context of mobile laser scanning. Equation~\ref{eq:lc1} thus becomes
\begin{align}
    \tilde \Gamma^n_{b,t_1} R^b_L v^L_1 - \Gamma^n_{b,t_2} R^b_L v^L_2 & = \nonumber \\
    R^b_L v^L_1 - \underset{\displaystyle \Gamma^{b,t_1}_{b,t_2}}{\underbrace{\left[ \tilde \Gamma^n_{b,t_1}\right]^{-1} \tilde \Gamma^n_{b,t_2}}} R^b_L v^L_2 & = 0 + \xi,
    \label{eq:lc2}
\end{align}
where in the second line it has been rearranged to show that the introduced measurement model only constrains the \emph{relative} body frame position and orientation between $t_1$ and $t_2$.


\textbf{\subsection{Dynamic Network structure}}

\begin{figure*}[!ht]
\centering
\includegraphics[width=14cm]{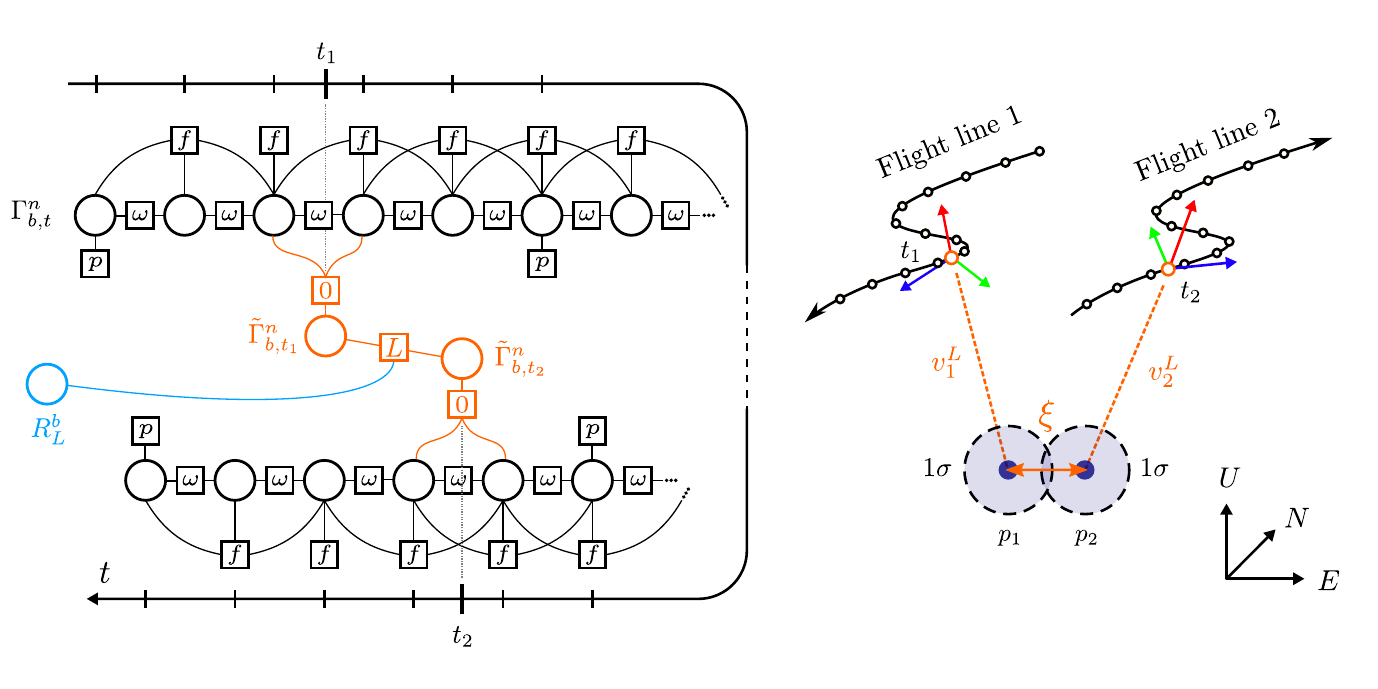}
\caption{On the left, the simplified structure of the DN hyper-graph in which only one correspondence edge $L$ has been shown (in orange) with the optional LiDAR boresight node (in blue). The nodes and edges required to handle the stream of GNSS and inertial measurements are shown in black. We have omitted the inertial sensor bias nodes for simplicity. The corresponding three-dimensional depiction is shown on the right.  }
\label{fig:corr_in_dn} 
\end{figure*}

The resulting DN structure considered in this work is presented in Fig.~\ref{fig:corr_in_dn}, where the circles represent unknowns, and the squares represent measurement hyper-edges, respectively. Five types of edges are considered:
\begin{enumerate}
    \item $p$: GNSS position measurements,
    \item $\omega$ and $f$: angular velocity and specific force edges. Here, IMU pre-integration is employed in order to reduce the overall number of unknowns in the graph without loss of information. This technique is very popular in visual-inertial odometry and SLAM, see for example~\citep{forster2016manifold}. The exact formulation is presented in~\citep{cucci2019onraw}. The inertial sensor biases and the related stochastic models are handled as in~\citep{cucci2017general}.
    \item{$0$: \emph{zero}-observation edges (edges for which no actual sensor measurement exists) that constrain $\tilde \Gamma^n_{b,\cdot}$ to lie on the geodetic in $\text{SE}(3)$ connecting adjacent, IMU synchronous, poses. 
    Please refer to~\citep[Section 4.2]{cucci_rawbundle_2017}. }
    \item{$L$: LiDAR correspondences, the error model being defined in Equation~\ref{eq:lc2}}.
\end{enumerate}

For further details, the reader is invited to refer to the reference publications and to the released source code, see Section~\ref{sec:refimpl} below.

\section{Design of experiments}
In this section, we describe the particular setup of the data acquisition and details of the data processing. We also present experiments that were conducted with the aim of highlighting the limitations of trajectory determination using lightweight IMUs and Kalman optimal smoothers as well as their impact on the georeferencing of the measured point cloud. Moreover, we seek to  demonstrate the potential benefit of employing the proposed point--to--point LiDAR correspondences together with raw navigation data within the Dynamic Network.

\subsection{Data acquisition}
\begin{figure}[h!]
\centering
\includegraphics[width = 8cm]{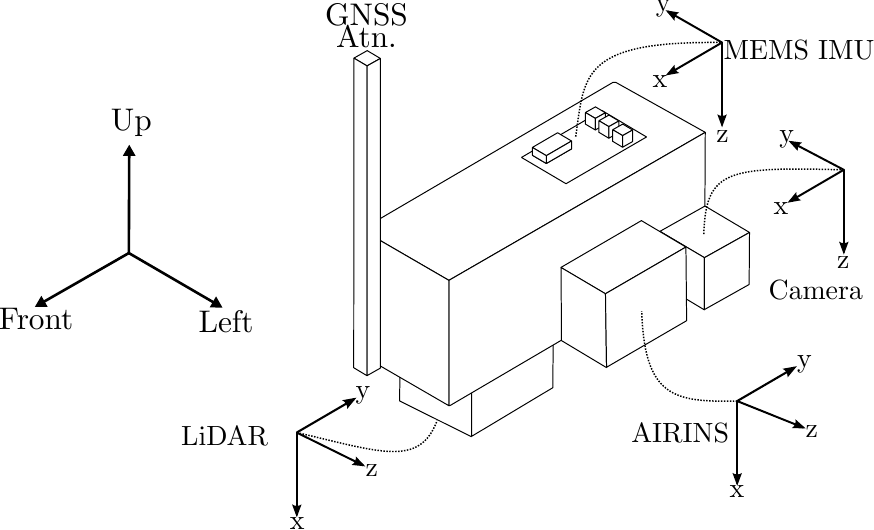}
\caption{\label{fig:mounting}Illustration of the mounting used to capture the data}
\end{figure}
For the experiments, we use the airborne dataset presented in \citep{ValletPuck2020}. There, reference and lower-cost sensors (LiDARs, IMUs and cameras) were rigidly mounted together on a vibration dampened assembly and installed in a helicopter. The common mounting provides the same flight conditions for all sensors (temperature, dynamics, height of flight, GNSS availability). In our experiments, we consider only part of those sensors, namely the reference LiDAR and two (reference and low-cost) IMUs. The reference LiDAR is a medium range VQ-480 (Riegl). The reference IMU is a navigation grade AIRINS (iXblue) while the small MEMS-IMU is a Navchip v1 (Thales), the performance of which is similar to popular commercial UAV-grade INSs such as the APX15 (Applanix) (see \cite{ValletPuck2020} or \cite{ClausenPlans2020} for a detailed evaluation). A scheme of the experimental mounting is shown in Fig. \ref{fig:mounting} and specifications of all sensors are presented in Table~\ref{tab:sensors}. \\
\begin{table}[h]
\centering
\resizebox{8.5cm}{!}{
\begin{tabular}{ccccc} 
\hline
\textbf{IMUs}  & \begin{tabular}[c]{@{}c@{}}Gyro drift\\{[}°/h]\end{tabular}        & \begin{tabular}[c]{@{}c@{}}Accel. drift\\{[}mg]\end{tabular}        & \begin{tabular}[c]{@{}c@{}}RP/Y acc.\\(PPK) [°]\end{tabular} & \begin{tabular}[c]{@{}c@{}}XYZ acc.\\(PPK) [m]\end{tabular}  \\ 
\hline
AIRINS         & 0.01                                                               & $<$0.25                                                                & 0.002/0.005                                                  & ref                                                         \\
Navchip          & 20                                                                 & 2                                                                   & 0.03/0.18                                                    & $<$0.03                                                         \\ 
\hline\hline
\textbf{LiDAR} & \begin{tabular}[c]{@{}c@{}}Beam\\divergence\\{[}mrad]\end{tabular} & \begin{tabular}[c]{@{}c@{}}Point meas.\\rate \\{[}kHz]\end{tabular} & \begin{tabular}[c]{@{}c@{}}Scan\\rate\\{[}Hz]\end{tabular}   & \begin{tabular}[c]{@{}c@{}}Return\\mode\end{tabular}         \\ 
\hline
VQ-480          & 0.27                                                               & 200                                                                 & 100                                                          & multi                                                        \\
\hline
\end{tabular}}
\caption{\label{tab:sensors}Manufacturer or confirmed specifications \citep{ValletPuck2020,ClausenPlans2020}}
\end{table}

\begin{figure}[h!] 
\centering 
\includegraphics[width=8cm]{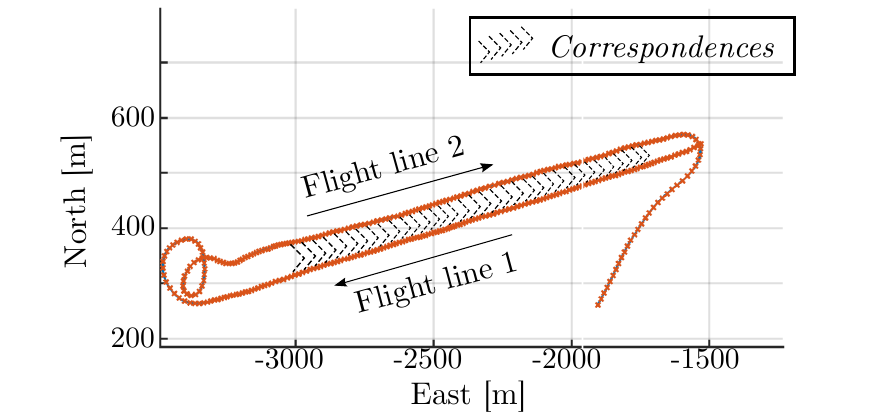} 
\caption{\label{fig:trajectory} Flight lines and position of the overlapping areas where correspondences were established.}
\end{figure}

During the flight, the helicopter flew profiles close to the ground to mimic a typical UAV flying altitude and at a speed similar to that of a small multi-copter (around 12 m/s) over an area featuring various terrain types including urban and rural areas, forest, croplands, roads, railroads and power lines. We focus on two successive flight lines depicted in Fig.\ref{fig:trajectory} that are approximately 2 km long, for a total flight time of around 6 minutes. The overlapping sections where correspondences can be established are depicted in the same figure. The swath width of each flight line is about 180 m and the side overlap is close to 40\%. It is important to mention that the reference LiDAR is not a small, UAV grade sensor and thus generates a higher point cloud density when flown at a typical UAV flight altitude. To mitigate this aspect, the helicopter was flown at 230 meters above ground level (AGL) while small UAV-based LiDARs are generally flown below 180 m AGL (see for example miniVUX \citep{minivux}). This results in point clouds generated with point densities varying between 35 pts/m$^2$ and 50 pts/m$^2$ on bare ground. Similar point cloud densities could be obtained by flying a UAV at a slightly lower speed (e.g., 10 m/s) and between 100 m and 160 m AGL. This density corresponds to a point cloud GSD (average distance between one point and its closest neighbor) of between 10 cm and 20 cm. 
\\
Thanks to this particular setup, it is possible to obtain a reference dataset that is based on a carrier phase differential post-processed (PPK) trajectory generated with the AIRINS coupled with the medium range LiDAR. The trajectory integrated with this kind of IMU shows attitude errors typically one order of magnitude lower than those based on MEMS-IMUs ($<$0.003° see \cite{ValletPuck2020}). This ensures low errors in the subsequent georeferenced point cloud ($\leq$ 5 cm). This dataset was considered as the ground-truth reference for our experiments due to its high accuracy.
\\
A ``UAV-grade'' data set was also generated with the setup described above. This was accomplished by employing the small MEMS-IMU (as is used on UAVs) to estimate the trajectory used for the subsequent georeferencing of laser observations. Albeit based on a larger LiDAR, the newly introduced line of UAV-targeted laser scanners from the same manufacturer (miniVUX) is reported to have ranging errors within PPK noise level (i.e. $<$ 2-3 cm) at up to $\sim$250 m AGL \citep{minivux}. Further, as the attitude quality based on pre-calibrated NavChip data is similar to that of APX15 \citep{ValletPuck2020} used on such scanners \citep{minivux}, we obtain a representative ``UAV-grade'' point cloud in terms of geometrical accuracy through applying the standard georeferencing procedure. This quality is then compared to that obtained via the proposed methodology.

\subsection{Data processing}
In all cases the absolute position and velocity observations are obtained via PPK from Javad multi-constellation and multi-frequency receivers (rover and master) using GrafNav (Novatel) software. Lever-arms between all sensors on the system are determined within 1 cm following the laboratory calibration approach of \cite{Vallet2004}.
The processing steps for the standard approach and the Dynamic Network approach with correspondences are summarized in Fig.~\ref{fig:std_pipeline} and Fig.~\ref{fig:pipeline}, respectively. For the sake of clarity we refer to the standard trajectory estimation approach based on the optimal Kalman Smoother as ``KS`` and to the Dynamic Network approach as``DN'' or ``DN+C'', the latter being the DN which considers LiDAR correspondences (``C'') as additional observations.

In the KS approach, we generate trajectories via a Kalman Smoother to integrate IMU readings in a loosely coupled manner with vehicle PPK position and velocity. We use the APPS software (iXblue) when integrating AIRINS reference data and Posproc (Applanix) with internally designed filters for the MEMS-IMU data. The LiDAR's boresight matrix is estimated with Riprocess (Riegl), which produces practically the same value as estimated with LIBOR \citep{LIBOR}. These parameters are estimated using separate flight data and reused for every dataset (with the exception of Case 3). The boresight between the AIRINS and the MEMS-IMU is determined as a mean value that minimizes the attitude differences between the two trajectories obtained by applying the Kalman smoother to data from both IMUs over almost 1 hour. While the boresight determined this way may be slightly incorrect due to residual misalignments in the trajectory determined with the MEMS-IMU, it ensures the best results for all ``KS'' trajectories in later comparisons. We thus stay on the ``safe side'' when comparing reference methodologies with the proposed one.
Finally, the point cloud is georeferenced based on the previously estimated trajectory and mounting parameters using an in-house software, LIEO \citep{LIEO}.

In the DN+C approach (Fig.~\ref{fig:pipeline}), the two first steps are common to the KS approach, where an approximate trajectory and subsequent point cloud are obtained using a Kalman Smoother (or using a Dynamic Network without LiDAR correspondences). We then establish correspondences as detailed in Sec. \ref{sec:correspondences}. For the preprocessing, a size of 50 by 50 meters was used to tile the point clouds. For the ISS key point detection method, point neighborhoods are extracted with a support radius of 1 meter and key points are selected using an eigen value ratio of 0.5, which enables the extraction of key-points from a variety of elements (man made and vegetation). For the LCD key-point description method, point neighborhoods are extracted with a support radius of 2 meters to include typical terrain features (e.g., building sections, cars, etc.) that lie within the neighborhood of the described points. A threshold of 25 cm (1.2 - 2.5 x GSD) was selected for the RANSAC algorithm. This threshold is important to perform non rigid matching and has been selected in consideration of the dynamic of the flight, the GSD and the size of the point cloud tiles. As will be shown later, such a threshold is suitable for the dynamic of standard UAV flights as well as for challenging scenarios such as during GNSS outages or in the case of miscalibrated boresight matricies, where trajectory estimation is less accurate and larger misalignments occur. Once 3D point--to--point LiDAR correspondences are generated, they are processed in a single pass of the DN along with raw inertial observations (angular rate and specific force), and GNSS positions. After such ``DN+C'' adjustment, the point cloud registration is refined based on the adjusted trajectory in LIEO for comparison with the reference. As noted above, trajectories can also be estimated in the Dynamic Network without using LiDAR correspondences (``DN'').
We investigate the proposed methodology within the following use cases and scenarios.

\subsection{Case 1: Optimal GNSS reception}

The trajectories described in this section are generated under optimal GNSS signal reception to serve as reference trajectories and point clouds obtained in ideal flight conditions. We define five scenarios:
\begin{itemize}
    \item Trajectory 0$\cdot$R: reference trajectory, employing navigation grade INS and KS approach. 
    All other trajectories and point clouds will be compared to this '' ground-truth`` dataset to estimate their respective accuracy.
    \item Trajectory 1$\cdot$KS: employing MEMS-IMU (as that employed in UAVs) and KS approach. 
    \item Trajectory 2$\cdot$DNC: employing MEMS-IMU (as that employed in UAVs) and DN+C approach. This dataset is obtained following the methodology depicted in Fig.\ref{fig:pipeline}. We establish correspondences on the UAV grade point cloud from trajectory $1$ in the area depicted in Fig.\ref{fig:trajectory}. In total, $\sim62'000$ correspondences are generated and used to constrain the adjusted trajectory, which is further used to refine the registration of the point cloud. 
    \item Trajectory 3$\cdot$DN: employing MEMS-IMU (as that employed in UAVs) and DN approach (i.e.,without correspondences). We produce this trajectory to be able to assess separately the impact of the LiDAR correspondences and the impact of the DN itself compared to a Kalman Smoother. 
    \item Trajectory 4$\cdot$DNC (from DN): employing the same procedure as trajectory 2$\cdot$DNC except that correspondences are generated from point clouds from trajectory 3$\cdot$DN instead of trajectory 1$\cdot$KS. The purpose of this setup is to show that the proposed approach does not rely on standard KS to generate the approximate trajectory and point cloud and that the DN can be used at both trajectory estimation steps. 
\end{itemize}
\subsection{Case 2: Quantitative impact of correspondences}

In this case we aim to assess the performance of the proposed procedure with a sparser set of LiDAR correspondences. This is to consider mapping an area with fewer geometric features to match, for example in a built up area with bare ground and without vegetation. To do so, we down-sample the set of point--to--popint LiDAR correspondences used in T. 2$\cdot$DNC from $50$\% to $0.1$\% of its initial size. We finally estimate a trajectory and subsequent point clouds for each previously described scenario and estimate the respective attitude and georeferencing errors.

\subsection{Case 3: Boresight self-calibration}

This scenario aims at assessing the ability of the method to calibrate the LiDAR boresight, $R^b_L$, ``in one go'' along with the trajectory determination, despite the sub-optimal geometry (two opposite flight lines at the same altitude). The following trajectories and subsequent point clouds will be compared:
\begin{itemize}
    \item Trajectory 5$\cdot$KSB: employing the same technique as trajectory 1$\cdot$KS, but without using the pre-calibrated boresight. To do so, each boresight angle is set to $0^\circ$ (unknown) when registering the point cloud in LIEO.
    \item Trajectory 6$\cdot$DNB: employing the same technique as trajectory 2$\cdot$DNC, but without using the pre-calibrated boresight. The boresight is considered as a supplementary static parameter to estimate by the DN with an initial value of $0^\circ$. Please see Fig.~\ref{fig:corr_in_dn}, in blue.
\end{itemize}

\subsection{Case 4: GNSS outage}

In this case we are interested to assess the ability of the proposed method (DN+C) to generate trajectories and point clouds of a higher accuracy under challenging scenarios such as during a GNSS outage.
\begin{itemize}
    \item Trajectory 7$\cdot$KSo: employing the same technique as trajectory 1$\cdot$KS, but with a simulated GNSS outage impacting the two flight lines during 60 seconds each (2 minutes in total). This corresponds roughly to a 750 m long area without GNSS signal reception  that could occur for example when operating in obstructed corridors, urban canyons or due to radio signal interference, or possibly jamming.
    \item Trajectory 8$\cdot$DNCo: employing the same technique as trajectory 2$\cdot$DNC, but with the same simulated GNSS denied area as trajectory 7$\cdot$KSo.
    \item Trajectory 9$\cdot$DNCo1: employing the same technique as trajectory 2$\cdot$DNC, but with a GNSS outage impacting only one of the two flight lines. This is to simulate a perturbation due to close physical obstruction (building, mountain) or a temporal electromagnetic disturbance for example. 
    This scenario is considerably different from trajectory 8$\cdot$DNCo because for all correspondences, only one of the 2 points is affected by the outage.
    \item Trajectory 10$\cdot$DNo: employing the same technique as trajectory 3$\cdot$DN, but with the simulated GNSS outage of 60 second over the two flight lines. We generate this trajectory such that we can observe separately the behavior of the correspondences and the DN estimated trajectory during outages.
\end{itemize}

Table~\ref{tab:case_summary} summarizes the different scenarios of trajectory estimation and their respective parameters.

\begin{table}[h!]
\small
\centering
\begin{tabular}{|l|c|c|c|c|} 
\hline
Traj. \# & \begin{tabular}[c]{@{}c@{}}Est. \\Method\end{tabular} & \begin{tabular}[c]{@{}c@{}}\% \\Corresp.\end{tabular} & \begin{tabular}[c]{@{}c@{}}Known\\bore.\end{tabular} & \begin{tabular}[c]{@{}c@{}}GPS \\outage\end{tabular} \\ 
\hline
T 1$\cdot$KS & KS & n.a. & \ding{51} & \ding{53} \\ 
\hdashline[1pt/1pt]
T 2$\cdot$DNC & DN & 100\% & \ding{51} & \ding{53} \\ 
\hdashline[1pt/1pt]
T 3$\cdot$DN & DN & 0\% & \ding{51} & \ding{53} \\ 
\hdashline[1pt/1pt]
\begin{tabular}[c]{@{}c@{}}T 4$\cdot$DNC\\(from dn)\end{tabular} & DN & 100\% & \ding{51} & \ding{53} \\ 
\hline
T 5$\cdot$KSB & KS & n.a. & \ding{53} & \ding{53} \\ 
\hdashline[1pt/1pt]
T 6$\cdot$DNCB & DN & 100\% & \ding{53} & \ding{53} \\ 
\hline
T 7$\cdot$KSo & KS & n.a. & \ding{51} & \ding{51} \\ 
\hdashline[1pt/1pt]
T 8$\cdot$DNCo & DN & 100\% & \ding{51} & \ding{51} \\ 
\hdashline[1pt/1pt]
T 9$\cdot$DNCo1 & DN & 100\% & \ding{51} & \begin{tabular}[c]{@{}c@{}}\ding{51} \\(single\\flight line)\end{tabular} \\ 
\hdashline[1pt/1pt]
T 10$\cdot$DNo & DN & 0\% & \ding{51} & \ding{51} \\
\hline
\end{tabular}
\caption{\label{tab:case_summary}Overview of evaluated trajectories and their parameters}
\label{tab:trajs}
\end{table}

\subsection{Reference implementation}
\label{sec:refimpl}

We release a reference implementation of the DN solver employed in this work. The code to compute all ``DN'' trajectories referenced in Table~\ref{tab:trajs}, along with the necessary data, is available at\footnote{\url{www.github.com/XXX} The source code will be released with the final version of this work.}.  This implementation relies on the ROAMFREE sensor fusion framework \citep{ROAMFREE1,ROAMFREE2}, a general purpose, open-source Dynamic Network solver that relies on the widely employed $\text{g}^2\text{o}$ least-squares solver~ \citep{Kummerle2011} and sparse linear algebra libraries \citep{sparse_alg}, allowing the adjustment of, for example, trajectory \hbox{2-DNC}, within (5 to 10) minutes on a standard desktop computer.

\section{Experimental results}

The dataset generated with navigation grade INS/PPK and precise airborne scanner (T. 0$\cdot$R) serves as the ground-truth reference with respect to which all other generated datasets are compared. It is important to keep in mind that the same LiDAR measurements are used in every dataset but the IMUs and methods used in georeferencing vary. This means that point clouds registered using different trajectories contain exactly the same number of points and that, for each point in these point clouds, the corresponding point in the ground-truth can be (easily) recovered. Thus, the ``mapping'' quality can be evaluated by comparing the difference between the coordinates of each point with the reference.
\subsection{Case 1: Optimal GNSS reception}

\begin{figure}[!ht] 
\centering 
\includegraphics[width=8cm]{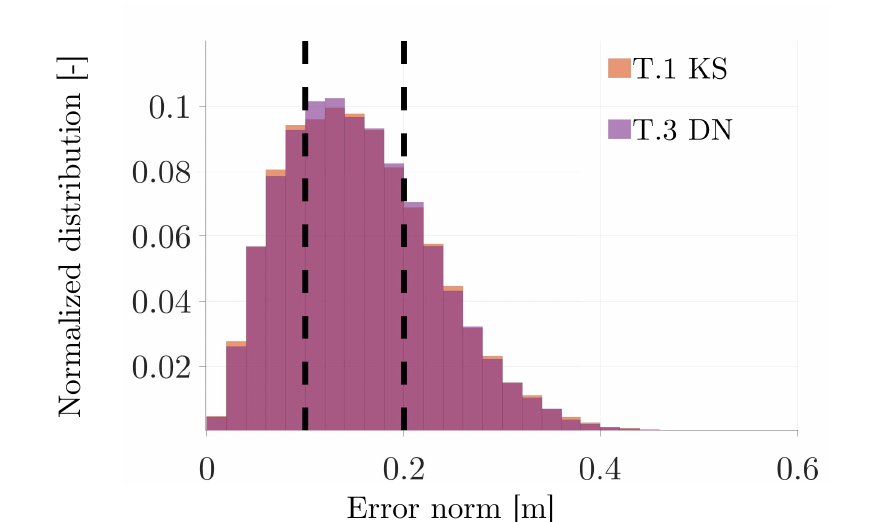}
\caption{\label{fig:corres_dist} Correspondence normalized error distribution. The actual interval of point cloud GSD is depicted in dashed black lines.}
\end{figure}

\textbf{Correspondences:} Correspondences are established following the methodology from Section \ref{sec:correspondences} (Fig.~\ref{fig:matching}) over an area depicted in Fig.~\ref{fig:trajectory} independently on two datasets. First, on the UAV-grade trajectory T.1$\cdot$KS (to be used in T. 2$\cdot$DNC) and second, on the T. 3$\cdot$DN (to be used in T. 4$\cdot$DNC (from DN)).\\ 
In both cases, the overlapping area is split into 91 tiles of 50 by 50 meters. Before filtration, between 200 and 1200 correspondences are retrieved per tile in built-up zones, while for tiles with vegetation between 1'000 and 6'000 correspondences are retrieved. This is due to the higher number of key points identified by ISS in vegetation. Overall, $\sim$50'000 correspondences are kept after filtration. We trace the two correspondence sets in the ground-truth point cloud to estimate the correspondence quality and estimate the error distribution as shown in Fig.~\ref{fig:corres_dist}.

First of all, we can see that the error distribution is almost identical in the two cases which means that the methodology for establishing the correspondences is not particularly dependent on method for the initial trajectory estimation.
We also observe that, after filtration, most of the correspondences have an error between 10 cm to 20 cm. This is mostly due to the GSD of the point cloud and the limited characterization of LCD that cannot efficiently separate points that are very close when projected in feature space (due to high similarity of their neighborhood). In other words, the mean fit is directly related to the point cloud GSD (in our case 0.1--0.2 m). Although increasing the sampling density could possibly allow detecting closer correspondences and thus obtaining stronger constraints between trajectories, we will demonstrate later that it is not necessary. In both cases, the mean error of retrieved correspondences is 15.6 cm with a 1$\sigma$ dispersion of 7.6 cm. As 15 cm corresponds to the average point cloud GSD, this confirms the suitability of choosing the GSD as a 1$\sigma$ value to weight correspondence constraints in the Dynamic Network (Fig.~\ref{fig:corr_in_dn}, right and Equation~\ref{eq:lc1}).

\begin{table*}
\centering
\begin{tabular}{lc|ccccc|ccc|} 
\cline{3-10}
 &                                                                                                & \multicolumn{5}{c|}{\textbf{Position Error [m]}}                                                                                            & \multicolumn{3}{c|}{\textbf{Attitude Error [°]}}                                     \\ 
\hline
\multicolumn{2}{|c|}{\multirow{2}{*}{\begin{tabular}[c]{@{}c@{}}\textbf{Traj.}\\\end{tabular}}}   & \multicolumn{1}{c|}{East} & \multicolumn{1}{c|}{North} & \multicolumn{1}{c|}{Up}   & \multicolumn{2}{c|}{Norm}                              & \multicolumn{1}{c|}{Roll} & \multicolumn{1}{c|}{Pitch} & Yaw                         \\ 
\cline{3-10}
\multicolumn{2}{|c|}{}                                                                            & \multicolumn{3}{c|}{RMSE}                                                          & \multicolumn{1}{c|}{Mean} & Std                        & \multicolumn{3}{c|}{RMSE}                                                            \\ 
\hline
\multicolumn{2}{|c|}{T.1 KS}                                                                      & 0.016                     & 0.016                      & 0.017                     & 0.023                     & 0.017                      & 0.037                     & 0.060                      & 0.190                       \\ 
\hline
\multicolumn{2}{|c|}{T.2 DNC}                                                                     & 0.008                     & 0.010                      & 0.011                     & 0.014                     & 0.010                      & 0.023                     & 0.021                      & 0.051                       \\ 
\hline
\multicolumn{2}{|c|}{T.3 DN}                                                                      & 0.007                     & 0.009                      & 0.014                     & 0.014                     & 0.010                      & 0.073                     & 0.053                      & 0.150                       \\ 
\hline
\multicolumn{2}{|c|}{\begin{tabular}[c]{@{}c@{}}T.4 DNC \\(from DN)\end{tabular}}                 & \multicolumn{1}{l}{0.008} & \multicolumn{1}{l}{0.010}  & \multicolumn{1}{l}{0.011} & \multicolumn{1}{l}{0.014} & \multicolumn{1}{l|}{0.009} & \multicolumn{1}{l}{0.023} & \multicolumn{1}{l}{0.021}  & \multicolumn{1}{l|}{0.046}  \\
\hline
\end{tabular}
\caption{\label{tab:traj_err}Position and attitude relative errors for trajectories n°1, 2, 3 and 4 with respect to the reference trajectory (T.0).}
\end{table*}

\begin{figure}[!ht] 
\centering 
\includegraphics[width = 8cm]{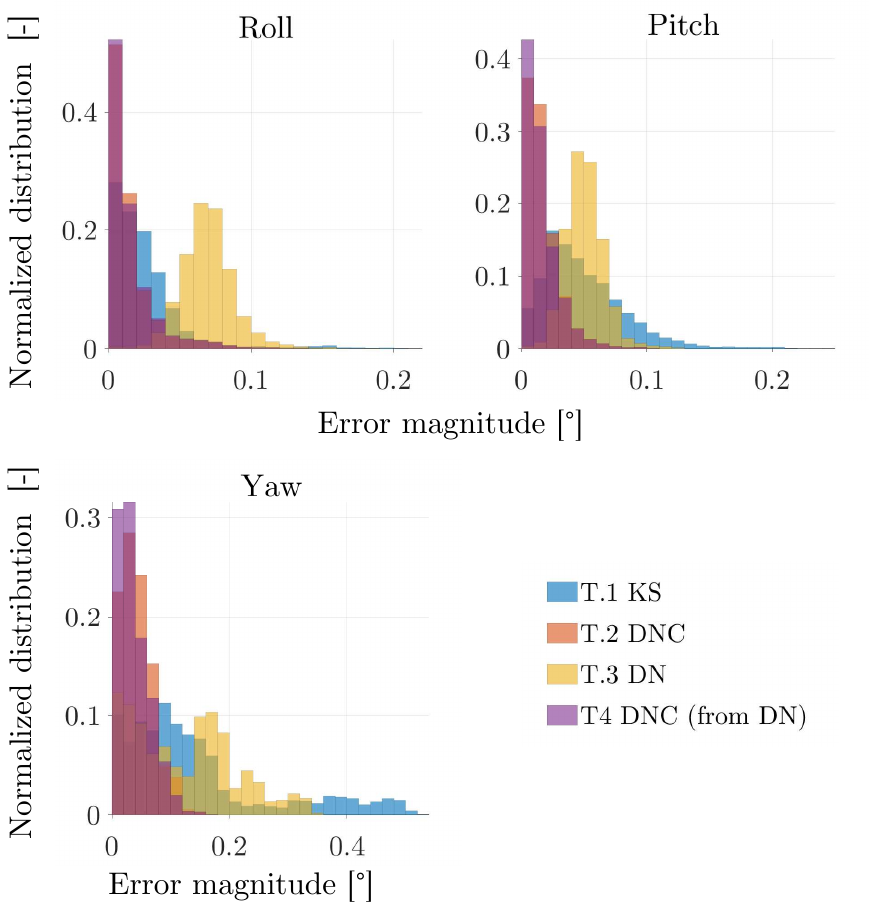}
\caption{\label{fig:rpy_hist} Normalized distribution of attitude errors in 4 trajectories.}
\end{figure}

\textbf{Trajectory:}
To estimate the accuracy of the trajectory, we compute the error in position (East, North, Up) and attitude (Roll, Pitch, Yaw), with respect to the reference. The impact of correspondences on the \textit{position} correction is expected to be limited within the nominal (no GNSS outage) scenario compared to the effect on \textit{attitude}. 
This is because, under such conditions, and with well calibrated system parameters, the main source of error affecting the point cloud georeferencing is the attitude obtained from small MEMS-IMUs \citep{Glennie2007a}. Indeed, the error in vehicle position with correctly determined differential carrier-phase ambiguities (PPK) is lower than the precision of correspondences.\\
We can see in Tab.~\ref{tab:traj_err} that the DN estimates the position slightly better than the Kalman Smoother with respect to the reference. Overall, the position mean error between T.1 and T.3 appears to be reduced by a factor of 1.7 (from 2.3 cm to 1.4 cm) but the respective differences are close to the reference precision. Hence, the inclusion of correspondences have practically negligible influence on the airborne position
for both T.2 and T.4 against T.3. This is somewhat expected since, as explained before, for the given GSD, the mean accuracy of correspondences is worse than vehicle position under PPK/INS. 
\\
However, 
Fig.~\ref{fig:rpy_hist} shows that using correspondences as input to the DN leads to a significant reduction of the error for all three attitude angles. Comparing the standard Kalman Smoother (T.1) to the proposed approach (T.2), the last three columns of Tab.\ref{tab:traj_err} show that the smallest correction occurs in the roll angle, with a 1.6x reduction of the RMSE (from 0.037° to 0.023°). This relatively small improvement is related to the fact that the absolute magnitude of the roll error is smaller compared to that of pitch and yaw, leaving less room for improvement. The improvements on pitch and yaw are more substantial. Pitch RMSE is reduced from 0.060° to 0.021° while Yaw RMSE is reduced from 0.190 to 0.051; a reduction factor of 3x and 3.7x respectively. Considering the fact that attitude quality improvement is proportional to the square (or even the cube) of the IMU size and weight (as well as cost), such improvement is substantial. Also, the ability to correct angles with larger initial errors is critical as they are responsible for most of the georeferencing errors (in the nominal mapping scenario with optimal GNSS signal reception). 
In this particular geometry, those larger angular errors cause the greatest discrepancies between the two flight lines and become observable with the inclusion of the 3D-LiDAR point--to--point correspondences. Pitch and yaw errors are observable along track (long overlapping area) while roll errors are visible across track (short overlapping area). Comparing T.2 and T.4 (i.e., the two DNC approaches based either on KS or DN to generate approximate trajectory and point clouds), we can see both reaching the same level of agreement with respect to the reference, despite their respective differences in initial trajectory errors and initial point cloud misalignment within the overlapping section. 
In this respect the error reduction is higher between both trajectories from the Dynamic Network (T.3 vs. T.4 as purple vs yellow in Fig.\ref{fig:rpy_hist}), which demonstrates that the use of correspondences provides the improvements in attitude estimation in this framework even with less complex inertial error modeling (e.g., with respect to Posproc) (which is the primary cause of attitude differences between T.1 and T.3 with respect to the reference).\\ 

\textbf{Point cloud:} As we have observed that the attitude estimation is greatly improved by the use of LiDAR correspondences within the Dynamic Network with respect to T.1, 
we expect the DN+C trajectory to deliver a point cloud of better quality than the one from the Kalman Smoother. To verify this, we estimate the point-wise georeferencing error by estimating for every point, the distance with respect to the corresponding point in the reference point cloud. 

\begin{figure}[!ht] 
\centering 
\includegraphics[width=8cm]{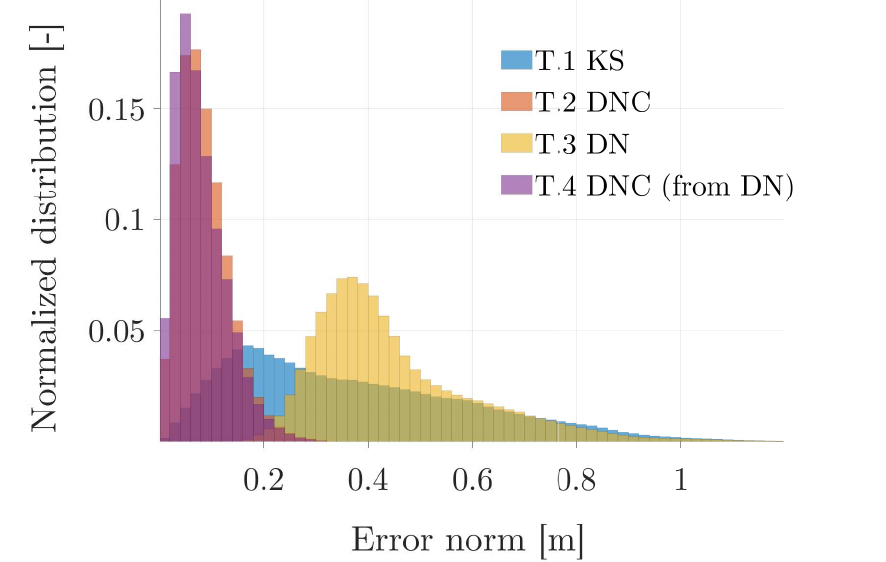}
\caption{\label{fig:pcd_hist}Normalized distribution of point cloud georeferencing error in magnitude for the four trajectories.}
\end{figure}

\begin{table}[!ht]
\centering
\begin{tabular}{|l|ccccc|} 
\hline
\multirow{2}{*}{\textbf{Traj.}}                                                   & \multicolumn{3}{c|}{RMSE [m]}                                                      & \multicolumn{2}{c|}{Norm [m]}                           \\ 
\cline{2-6}
                                                                                  & \multicolumn{1}{c|}{East} & \multicolumn{1}{l|}{North} & \multicolumn{1}{l|}{Up}   & \multicolumn{1}{c|}{Mean} & Std                         \\ 
\hline
T.1 KS                                                                            & 0.428                     & 0.120                      & 0.034                     & 0.379                     & 0.234                       \\ 
\hline
T.2 DNC                                                                           & 0.079                     & 0.058                      & 0.018                     & 0.087                     & 0.049                       \\ 
\hline
T.3 DN                                                                            & 0.310                     & 0.361                      & 0.094                     & 0.455                     & 0.166                       \\ 
\hline
\multicolumn{1}{|c|}{\begin{tabular}[c]{@{}c@{}}T.4 DNC \\(from DN)\end{tabular}} & \multicolumn{1}{l}{0.072} & \multicolumn{1}{l}{0.058}   & \multicolumn{1}{l}{0.018} & \multicolumn{1}{l}{0.080} & \multicolumn{1}{l|}{0.049}  \\
\hline
\end{tabular}
\caption{\label{tab:pcd_err}Point cloud georeferencing error in the 1st flight line}
\end{table}

The results are visible in Fig.~\ref{fig:pcd_hist}. On flight line 1, when using the standard KS approach (T.1 KS, in blue), the maximum georeferencing error is about 1.1 m and most of the points have errors between 15 cm and 50 cm (1$\cdot$GSD to 3$\cdot$GSD). These errors are significantly reduced when using the DNC methodology. Considering T.2$\cdot$DNC, most of the points are georeferenced within 5 cm to 15 cm (0.3$\cdot$GSD to 1$\cdot$GSD). The maximum georeferencing error is also greatly reduced, to 28 cm in this flight line. Comparing the two DNC cases (T.2 and T.4), georeferencing errors are almost identical due to similar trajectory improvements (see Tab.~\ref{tab:traj_err}). Finally, the errors are much larger using the Dynamic Network without correspondences (T.3 DN, yellow), which demonstrates that the reduction in georeferencing error is due to the use of correspondences. As shown in Tab.~\ref{tab:pcd_err}, the mean error using the standard Kalman Smoother approach (T.1,KS) is 38 cm and is decreased to 8 cm using the proposed approach (T.4), corresponding to a reduction by a factor of $\sim$5, from 2.5$\times$GSD to 0.6$\times$GSD. This improvement is mostly in planimetry, as shown by the RMSE per axis ( East/North columns in Tab.~\ref{tab:pcd_err}). Please note that the reference point cloud, obtained with a navigation grade IMU, is assumed to be accurate up to $5$ cm.

It is important to keep in mind that the magnitude of the georeferencing errors is proportional to the flight altitude, in our case 230 m AGL. Flying at a lower altitude and with less powerful and smaller scanners intended for use with UAVs, i.e., miniVUX1-3 \cite{minivux} at $\sim$150 m AGL, would allow the registration refinement to fully benefit from attitude improvement and the desired point cloud accuracy of 5 cm could be obtained.

\begin{figure}[!ht] 
\centering 
\includegraphics[width = 8cm]{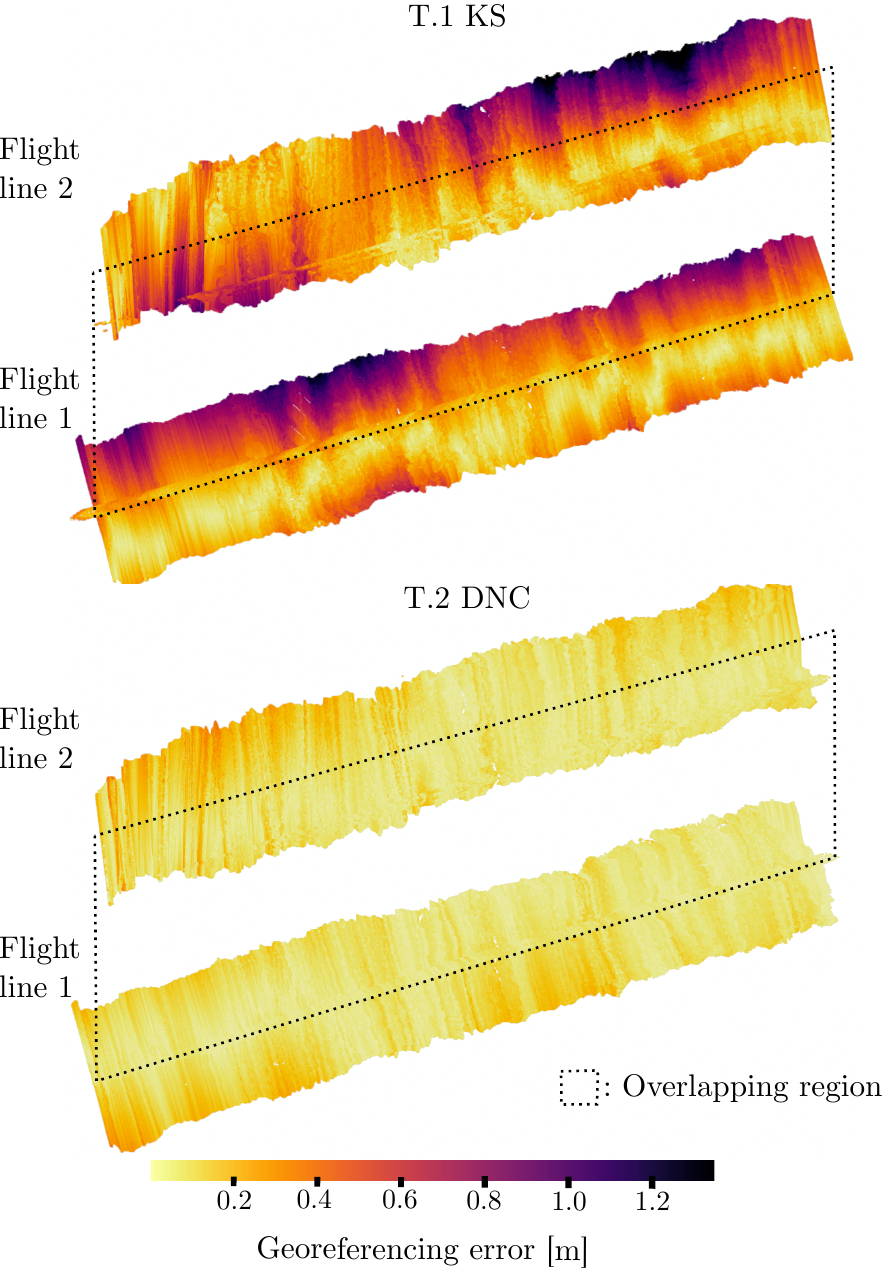}
\caption{\label{fig:pcd_err}Magnitude of error in points coordinates based on trajectory KS T.1 (top) and T.2$\cdot$DNC (bottom).}
\end{figure}

Fig.~\ref{fig:pcd_err} shows georeferencing errors at the scale of the point cloud by coloring each point according to its spatial distance with respect to the reference.
The point clouds in the upper part of this figure are georeferenced using T.1 that is based on the KS. As expected, larger errors are observed at the edges of the scan, due to the distance from the vehicle. While the error is below 30 cm at the center of the point cloud, it reaches 1.2 meters in the edges. The irregular patterns in the error are due to the orientation dynamic of the flight causing the rapid fluctuations in attitude error projection in the mapping frame (which are difficult, if not impossible to correct with ICP). The improvements are very clear when comparing these errors to the bottom part of the figure, which is based on DN+C trajectory (T.2). Here, the georeferencing errors are considerably reduced and most of the points have a georeferencing errors lower than 15 cm.\\

\textbf{Summary:}  This practical example demonstrates that within the nominal mapping scenario (i.e.,stable GNSS signal reception) using a small MEMS-IMU, incorporating point--to--point correspondences into the Dynamic Network significantly improves the quality of the computed trajectory and in particular the three attitude angles. In turn, this ameliorated estimation of the vehicle attitude results in residual point cloud registration errors that are lower than the GSD; a significant improvement compared to the direct georeferencing approach where the errors are $\sim$5$\times$ larger. 

\subsection{Case 2: Quantitative impact of correspondences}

\textbf{Correspondences:} In this situation we use the same set of correspondences as in T.4$\cdot$DNC (from DN), except that we down-sample them, or in other words we use a small sub-set of the original correspondence set in subsequent adjustments. Thus their quality is the same as depicted in Fig.~\ref{fig:corres_dist}, but fewer are used in the adjustment. In particular, we down-sample progressively from 50\% to 0.1\% of the original set of correspondences and estimate both trajectories with their respective point clouds for each step.

\begin{figure}[!ht] 
\centering 
\includegraphics[width=8cm]{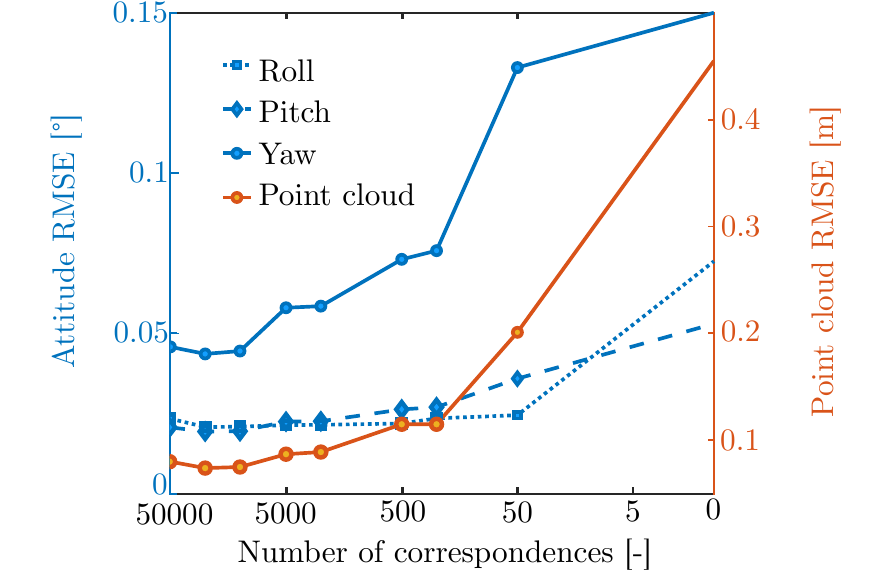}
\caption{\label{fig:subsample_effect}Effect of correspondences down-sampling on attitude estimation and subsequent point cloud regeneration.}
\end{figure}

\textbf{Impact:}
Fig.~\ref{fig:subsample_effect} illustrates the effect of down-sampling correspondences (x-axis, in logarithmic scale) on the attitude RMSE (left y-axis, in blue) and on the RMSE of the registered point cloud (right y-axis, in red). Yaw angle is the first angle affected by the down-sampling since its RMSE increase significantly when down-sampling more than 25\% (keeping $\sim$12'500 correspondences). On the other hand, the estimation of roll and pitch is less sensitive to down-sampling, their RMSE being stable up to down-sampling to 0.1\% (keeping only $\sim$50 correspondences) for roll and 0.5\% ($\sim$250 correspondences) for pitch. Due to the geometry of nadir ALS, laser beams are mostly projected along the vertical axis. While a perturbation in roll and pitch will impact more its vertical component, a perturbation in yaw affects more the horizontal component towards the swath extremities. 
For this reason, despite being more influenced by down-sampling as compared to other angles, yaw error has generally smaller effect on point cloud georeferencing than pitch and roll. The stability in roll and pitch estimation allows the point cloud to be accurately registerd even with strong correspondence down-sampling. In fact, the point cloud RMSE is kept below 10 cm when down-sampling at 5\% ($\sim$2500 correspondences), and is still remarkably stable up to a down sampling at 0.5\% ($\sim$ 250 correspondences) which results in an increase of the georeferencing RMSE by a factor of 1.4$\times$.\\

\textbf{Summary:} This example illustrates practically the ability of the proposed method to correctly adjust the trajectory even with a considerably smaller number of correspondences. This can be practical when scanning over an areas with limited distinctive features (e.g., crops with sparse and low vegetation) where establishing correspondences is more challenging.

\subsection{Case 3: Boresight self-calibration}
\begin{figure}[!ht] 
\centering 
\includegraphics[width=8cm]{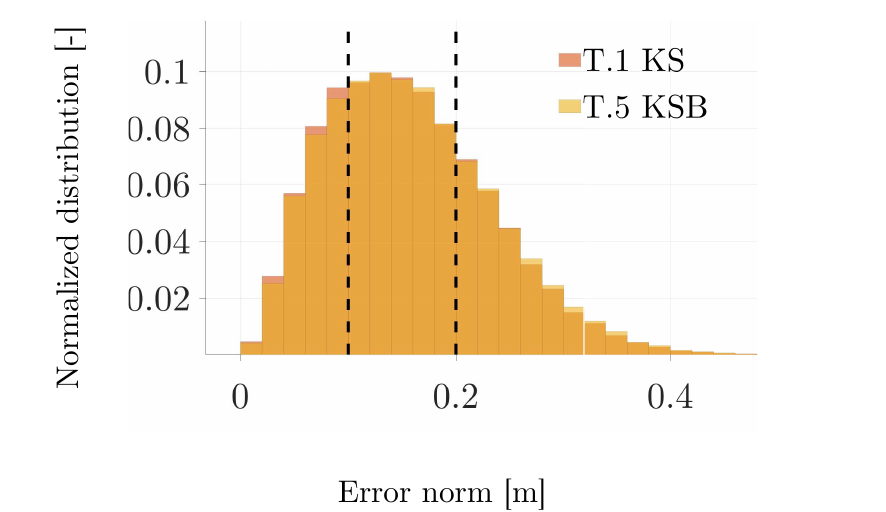}
\caption{\label{fig:corr_dist_bore} Correspondences' normalized error distribution. The actual interval of point cloud GSD is depicted in dashed black lines.}
\end{figure}

\begin{figure}[!ht] 
\centering 
\includegraphics[width=8cm]{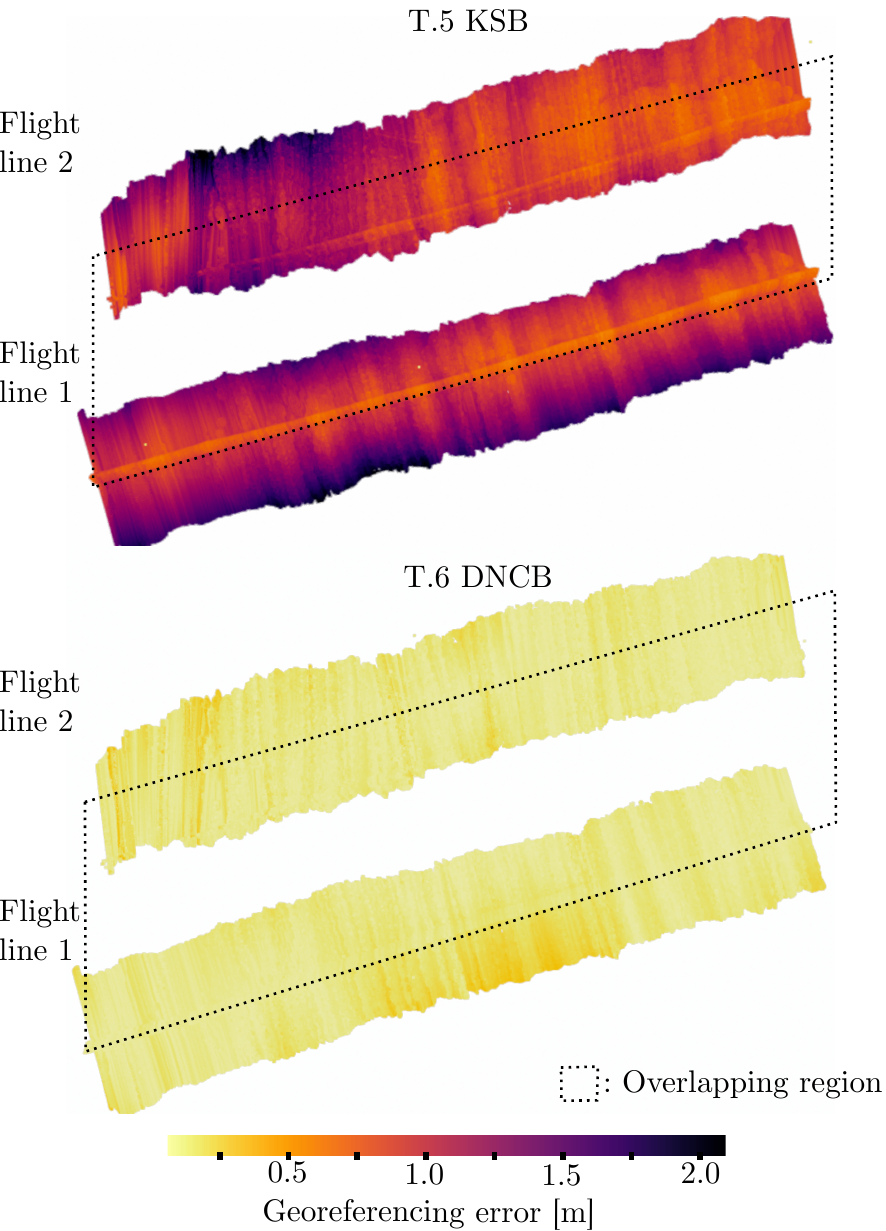}
\caption{\label{fig:map_hist_bore} Points  georeferencing  error with unknown (top) end recovered boresight (bottom) via constrained Dynamic Network.}
\end{figure}

\textbf{Correspondences:}
Correspondences are established on the point cloud registered using the KS trajectory with unknown (zero) boresight (T.5$\cdot$KSB). This dataset shows significantly higher misalignment between the point clouds due to the absence of the boresight parameters (e.g., between 1 and 3.5 meters in the overlapping section, in the top part of  Fig.~\ref{fig:map_hist_bore}). We first compare the agreement of established LiDAR correspondences with those obtained in the ``boresight-perfect'' dataset from T.1$\cdot$KS. As shown in Fig.~\ref{fig:corr_dist_bore}, the histogram of the two error distributions overlaps, indicating that the quality of the correspondences is almost identical in both cases. This confirms the translation invariance of the LCD descriptor and the ability of the method to establish sufficiently accurate correspondences even with a poor initial registration (i.e., between two strongly misaligned point clouds). 

The trajectory is then re-estimated along with the three LiDAR boresight angles by the Dynamic Network with the set recovered of correspondence. Both outputs (corrected trajectory and the estimated boresight) are finally used as input to LIEO to refine the registration of the point clouds.\\

\begin{figure}[!ht] 
\centering 
\includegraphics[width=8.5cm]{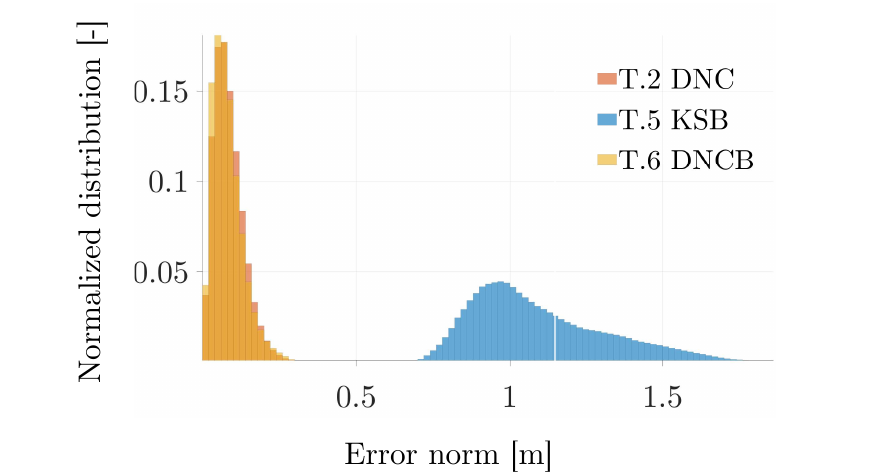}
\caption{\label{fig:pcd_hist_bore} Normalized distribution of point cloud georeferencing error in magnitude for trajectories T.2, T.5 and T.6.}
\end{figure}

\textbf{Point cloud:}
The quality of the point cloud registration is depicted in Fig.~\ref{fig:pcd_hist_bore} as a histogram of the georeferencing error (point-wise magnitude) for T.5$\cdot$KSB (blue) and T.6$\cdot$DNCB (yellow). For the sake of completeness, the same figure displays the error histogram for the reference point cloud based on T.2$\cdot$DNC (orange) as well. 

We first observe the effect of the absence of boresight parameters on the point cloud (T.5$\cdot$KSB, blue). This point cloud shows large georeferencing errors ranging from 0.7 m to 1.8 m and a mean error of 1.1 m (e.g., compared to 0.38 cm with the calibrated boresight, T.1$\cdot$KS).
On the other hand, we can see that, when coupled with correspondences, the Dynamic Network is able to estimate the boresight matrix as well as correct the systematic causes for the errors in the trajectory at the same time: as shown in bottom part of Fig.~\ref{fig:map_hist_bore}, the resulting point cloud based on T.6$\cdot$DNCB is well aligned. Indeed, the mean error is approximately 8 cm, which is about 13$\times$ smaller that that of the point cloud based on T.5$\cdot$KSB.

The boresight estimated by the Dynamic Network is slightly different from that obtained via the calibration flight using planar constraints (using Riprocess, Riegl), with values per angle of [-0.276 ; 0.049 ; 0.154] and [-0.213 ; 0.010 ; 0.191], respectively. This difference is mainly due to the fact that boresight angles and IMU attitude errors (caused by imperfect initialization of orientation that is, among others, related to the unknown biases of the inertial sensors) have similar effect on the trajectory, at least within this scenario. Since the Dynamic Network estimates those simultaneously, the change in boresight angles is compensated with IMU biases, leading to a resulting point cloud registration that is very similar to T.2$\cdot$DNC. \\

\textbf{Summary:} In this case we first showed practically that the proposed approach to estimated correspondences is robust to large misalignment in the approximate point cloud. Second, we demonstrated the ability of the ``DNC'' approach to correct for point cloud deformations due to unknown mounting parameters, in this case the three boresight angles, without any loss in accuracy losses in the final georeferenced point cloud.

\subsection{Case 4: GNSS outage}

\begin{figure}[!ht] 
\centering 
\includegraphics[width=8cm]{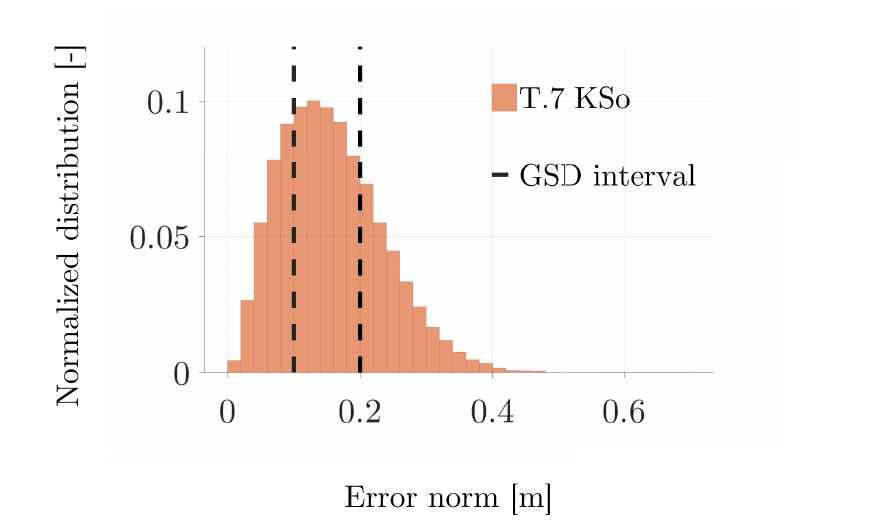}
\caption{\label{fig:corres_outage}Normalized error distribution of established correspondences. The actual interval of point cloud GSD is depicted by dashed black lines.}
\end{figure}

\textbf{Correspondences:} 
Correspondences are established on the point cloud generated from the KS trajectory with a relatively long (for a low-cost MEMS-IMU) GNSS outage of $60$ s (T.7$\cdot$KSo). We have seen in Sec.~5.1 that the quality of correspondences is not necessarily impacted when using a point cloud georeferenced with a highly sub-optimal trajectory. This is again verified for a trajectory affected by a 2 minute-long absence of GNSS position measurements. As can be seen in Fig.~\ref{fig:corres_outage}, the error distribution is almost identical to the cases discussed previously (see Fig.~\ref{fig:corres_dist}). After filtration, most of the correspondences have an error between 10 cm and 20 cm, while the mean error in the correspondences is 15.8 cm (compared to 15.6 cm in T.1$\cdot$KS and T.3$\cdot$DN) with 1$\sigma$ dispersion of 7.8 cm. 
Due to the similarity in correspondence quality, we use the same set of correspondences (originated from T.7$\cdot$KSo dataset, containing a double outage) in both T.8$\cdot$DNCo and T.9$\cdot$DNCo1.

\begin{figure}[!ht] 
\centering 
\includegraphics[]{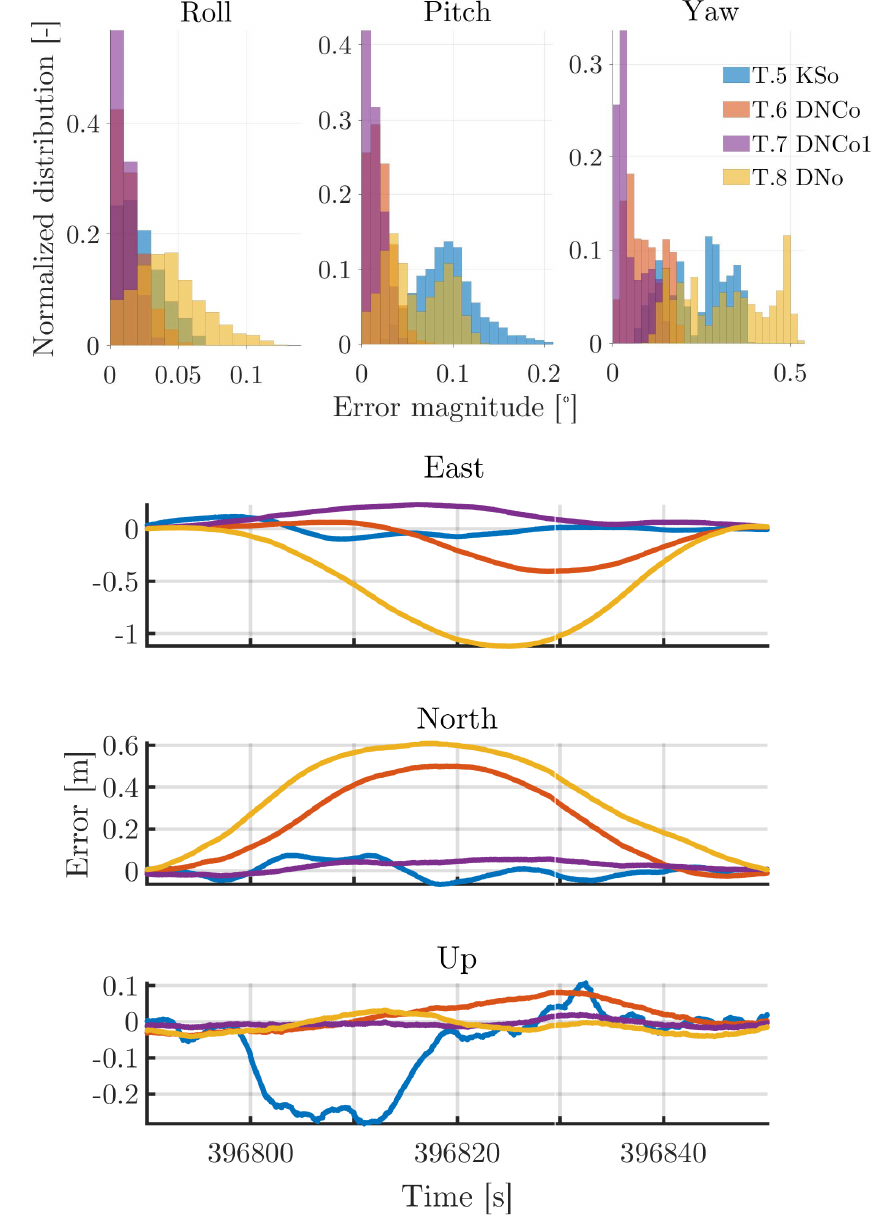}
\caption{\label{fig:outage}Errors in position and attitude during outage on line 1 (60 seconds) on 4 tested trajectories.}
\end{figure}

\textbf{Trajectory:}
Similarly to case 1, we estimate trajectory errors in terms of position and attitude. For a fair comparison, we estimate the errors only for the portion of outage within the first flight line, that is common to all four trajectories T.7 to T.10. The first 5 columns of Tab.~\ref{tab:out_error} as well as Fig.~\ref{fig:outage} (left) illustrate the effect of the outage on the position estimation.

\begin{table*}[h!]
\centering
\begin{tabular}{lc|ccccc|ccc|} 
\cline{3-10}
 &  & \multicolumn{5}{c|}{\textbf{Position Error [m]}} & \multicolumn{3}{c|}{\textbf{Attitude Error [°]}} \\ 
\hline
\multicolumn{2}{|c|}{\multirow{2}{*}{\textbf{Traj.}}} & \multicolumn{1}{c|}{East} & \multicolumn{1}{c|}{North} & \multicolumn{1}{c|}{Up} & \multicolumn{2}{c|}{Norm} & \multicolumn{1}{c|}{Roll} & \multicolumn{1}{c|}{Pitch} & Yaw \\ 
\cline{3-10}
\multicolumn{2}{|c|}{} & \multicolumn{3}{c|}{RMSE} & \multicolumn{1}{c|}{Mean} & Std & \multicolumn{3}{c|}{RMSE} \\ 
\hline
\multicolumn{2}{|l|}{T.7 KSo} & 0.058 & 0.036 & 0.126 & 0.109 & 0.094 & 0.027 & 0.099 & 0.251 \\ 
\hline
\multicolumn{2}{|l|}{T.8 DNCo} & 0.206 & 0.297 & 0.040 & 0.299 & 0.202 & 0.018 & 0.024 & 0.102 \\ 
\hline
\multicolumn{2}{|l|}{T.9 DNCo1} & 0.133 & 0.033 & 0.011 & 0.118 & 0.071 & 0.012 & 0.017 & 0.064 \\ 
\hline
\multicolumn{2}{|l|}{T.10 DNo} & 0.650 & 0.403 & 0.024 & 0.630 & 0.434 & 0.050 & 0.068 & 0.341 \\
\hline
\end{tabular}
\caption{\label{tab:out_error}Position and attitude errors in trajectories T.7 to T.10 during the outage on line 1 with respect to the reference.}
\end{table*}

We first observe that the Dynamic Network without correspondences (T10$\cdot$DNo), estimates airborne positions during a GNSS outage less precisely than the Kalman Smoother (T.7$\cdot$KSo). 
This is likely due to refined modelling of inertial errors that are considered within the Kalman Smoother, which results in a larger error state vector (i.e., including scale-factors for accelerometers and gyroscopes) as well as tuned ``stochastic models'' for the particular IMU used.  
Indeed, the experience with the Dynamic Network setup for this type of IMU is somewhat limited and the choice of models is limited.

Second, we notice that the impact of using correspondences depends on the outage scenario. When considering the outage on two flight lines (T.8$\cdot$DNCo), both points in each correspondence fall within the ``GNSS denied segment''. In such a situation, a constant shift in position estimation occurring along the same direction during the outages will not be detectable via correspondences because they are relative, as shown in Equation~\ref{eq:lc2}. This explains the limited efficiency of the correspondences to correct for trajectory position errors when both points are from a "GNSS denied area". 
Indeed, as shown in Tab.~\ref{tab:out_error}, the Dynamic Network with correspondences over the double outage (T.8$\cdot$DNCo) performs worse than the Kalman Smoother (T.7$\cdot$KSo), apart from the vertical component, that is improved by a factor 3.2. 

Finally, we compare the Kalman Smoother (T.7$\cdot$KSo) with the Dynamic Network (T.9$\cdot$DNCo1) facing a single outage. In this situation, only one of the flight lines is within the ``GNSS denied'' zone. Therefore a point originating from the trajectory that is not affected by the outage in each correspondence acts as an anchor for a point affected by the outage. This explains the superior correction capacity of the Dynamic Network in this scenario, which reaches comparable precision to the Kalman Smoother (error increased by a factor 2.5 along east axis but reduced by a factor 10 along up axis).  

By definition, the attitude is less impacted during a GNSS outage since the INS drifts slower in vehicle orientation. This is confirmed in Fig.~\ref{fig:outage} where incorporating correspondences within the Dynmaic Network during single and double GNSS outages results in a reduction of the attitude errors.
Particularly, in comparison to T. 5$\cdot$KSo, there is a reduction of attitude RMSE within T.9$\cdot$DNCo1,by a factor 2.3 and 5.8 in roll and pitch, respectively, and 3.9 in yaw. On the other hand, for the double outage T.8$\cdot$DNCo, the improvement is comparable to the KS scenario with uninterrupted GNSS signal reception, with reduction by a factor 1.5 in roll, 4.1 in pitch and  2.5 in yaw.\\



\textbf{Point cloud:}
We estimate the point-wise georeferencing error in the point clouds (with respect to  the reference cloud) for trajectories T.7 to T.10 . The axis-wise RMSE and mean georeferencing errors are displayed in Tab.~\ref{tab:pcd_err_outage}.

\begin{table}[h!]
\centering
\begin{tabular}{|l|ccccc|} 
\hline
\multirow{2}{*}{\textbf{Traj.}} & \multicolumn{3}{c|}{RMSE [m]} & \multicolumn{2}{c|}{Norm [m]} \\ 
\cline{2-6}
 & \multicolumn{1}{c|}{East} & \multicolumn{1}{c|}{North} & \multicolumn{1}{c|}{Up} & \multicolumn{1}{c|}{Mean} & Std \\ 
\hline
T.7 KSo & 0.591 & 0.121 & 0.112 & 0.523 & 0.321 \\ 
\hline
T.8 DNCo & 0.326 & 0.301 & 0.042 & 0.376 & 0.248 \\ 
\hline
T.9 DNCo1 & 0.158 & 0.055 & 0.019 & 0.130 & 0.107 \\ 
\hline
T.10 DNo & 1.089 & 0.312 & 0.099 & 0.977 & 0.582 \\
\hline
\end{tabular}
\caption{\label{tab:pcd_err_outage}Point clouds georeferencing error with respect to the 1st flight line.}
\end{table}

\begin{figure}[!ht] 
\centering 
\includegraphics[width=8cm]{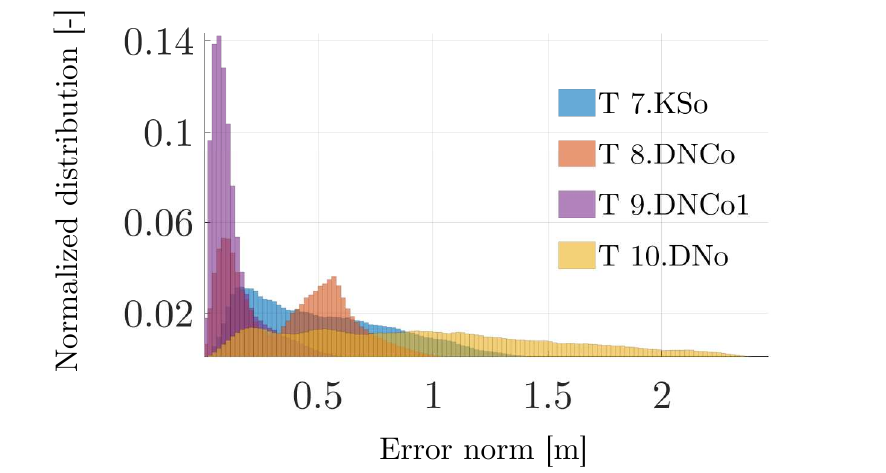}
\caption{\label{fig:pcd_err_outage}Normalized distribution of point cloud georeferencing error in magnitude for the 4 trajectories.}
\end{figure}

First we compare T.7$\cdot$KSo and T.8$\cdot$DNCo. Despite the relatively  good estimate of attitude, we have seen that T.8$\cdot$DNCo trajectory contains larger systematic effects in position during the outages than T.7$\cdot$KSo. This limits the improvement of the point cloud georeferencing, from 52 cm  (T.7$\cdot$KSo) to 38 cm (T8$\cdot$DNCo) within the GNSS denied area. 
Note that the error distribution (marked red in Fig.~\ref{fig:pcd_err_outage}) contains 2 peaks for T8$\cdot$DNCo. This is related to the residual error in trajectory estimation along East axis, that is correct only during the first half of the outage.

The improvements are more substantial when only one flight line is impacted by the GNSS signal obstruction (i.e., T.9$\cdot$DNCo1) as one side of the correspondences serves as a ``position anchor''). In this situation, 
the georeferencing error is reduced by a factor of 4 compared to T.7$\cdot$KSo -- from 52 cm to 13 cm ($\sim1\cdot$GSD). Finally, the Dynamic Network without correspondences manifests the largest georeferencing errors, a fact that is expected in the trajectory comparison (attributed to less optimal modeling of IMU) and confirming that correspondences are indeed responsible for the improvements in the point cloud regeneration within this methodology.  

\begin{figure}[!ht] 
\centering 
\includegraphics[width=8cm]{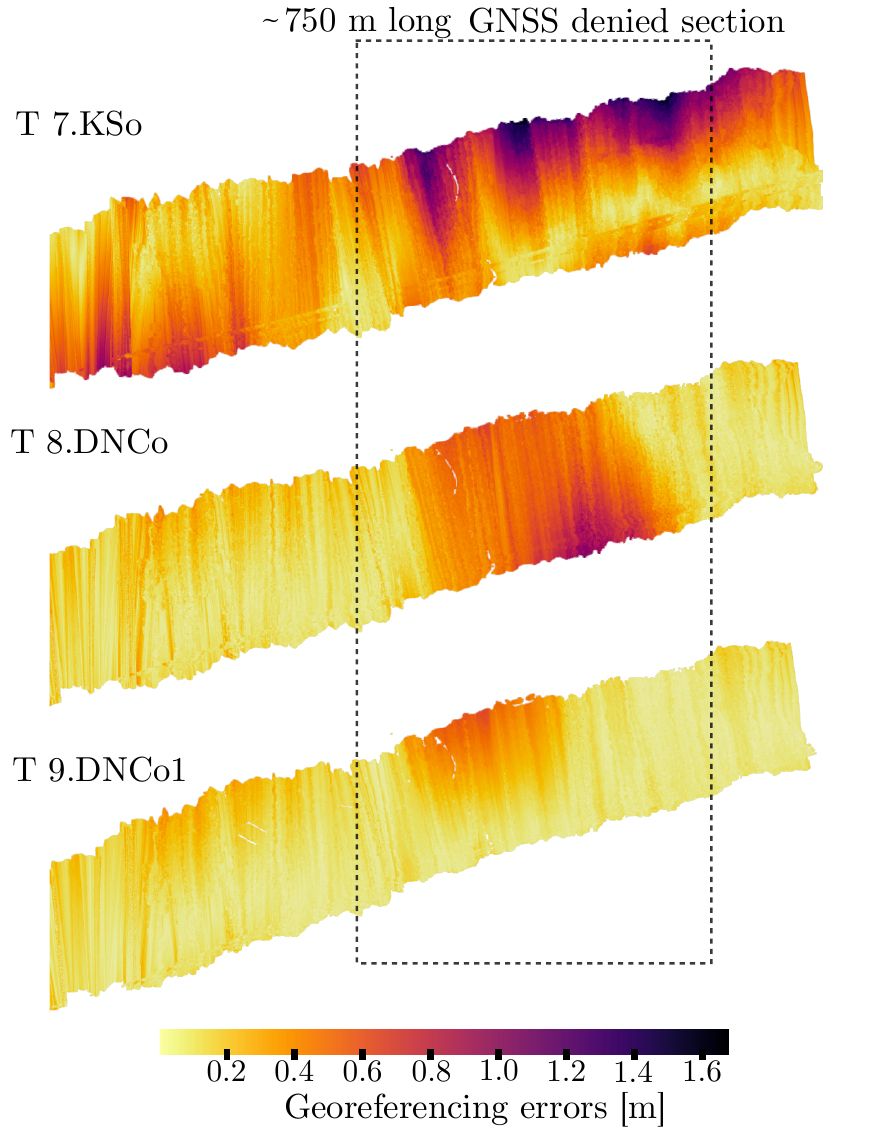}
\caption{\label{fig:pcd_outage_map}Normalized distribution of point cloud georeferencing error (magnitude) for the 4 trajectories.}
\end{figure}

The point cloud georeferencing error along the entire flight line is illustrated in Fig.~\ref{fig:pcd_outage_map}. Outside of the GNSS denied section, both T.8$\cdot$DNCo and T.9$\cdot$DCo1 show significantly lower errors than T.7$\cdot$KSo. Inside the GNSS denied section, we can observe that the error varies rapidly with larger deviations in T.7$\cdot$KSo as compared to T.8$\cdot$DNCo, where the error pattern is smoother but still remains widespread. This illustrates the difference between the impact of attitude errors (of which the projection on the point cloud coordinates is rapidly varying -- T.7$\cdot$KSo) and the position errors (that are slowly drifting, T.8$\cdot$DNCo) within the laser georeferencing.\\

 \textbf{Summary:} In this last case, we evaluated the performance of the proposed approach in the event of GNSS outages. We showed that the use correspondences within the propose approach produces a significant reduction in the trajectory attitude error as compared to the standard approach. The error in trajectory position is also significant reduced using the propposed methed when only one flight line is affected by an outage. This is due to the fact that one point in each correspondence serves as a position anchor to another point affected by the GNSS outage thereby enabling the correction of the part of trajectory within the GNSS denied segment. The situation is less clear when both flight lines experience an outage since and their respective ``deviations'' go in a similar direction and magnitude (in the mapping frame). Such a situation is not observable at the correspondence level to act as relative constraints. However, in our experiment, their inclusion still leads to a reduction of $\approx$30\% of the point cloud georeferencing error within the GNSS denied area. 
 
\section{Conclusion}
In this work we first used ``close-range'' airborne laser scanning (AGL $<$ 300 m) over terrain with different typology (low \& high vegetation, buildings, infrastructure)  to evaluate some existing methods in feature-based point--to--point matching. We combined the promising detector/descriptor with additional filtering to obtain robust LiDAR 3D point--to--point correspondences between overlapping areas. We integrated those as additional observations into a Dynamic Network which at its base considers GNSS positions and raw inertial data (specific forces and angular rates) in a single (iterative non-linear least-squares) adjustment. For a system with small MEMS-IMU (as used in UAVs), this ``tight'' sensor fusion resulted in an improved trajectory determination that in turn produced considerably better results in the point cloud registration process.\\

We investigated different scenarios related to the correspondence quality and quantity, lack of knowledge in system parameters, as well as limitations in GNSS signal reception, from which we can draw the following conclusions: 

\begin{itemize}
    \item The chosen detection, matching and filtering of over overlapping patches (50$\times$50 m)  resulted in a high number of correspondences irrespective of terrain type.
    
    \item The performance of correspondence quality was not affected by poor approximation of initial trajectory either due to a non-calibrated system or long GNSS outage. This indirectly indicates that the translation invariance properties of the employed descriptor were sufficient with respect to the size of patches used in the correspondence matching. Indeed, in all studied cases, the goodness of 3D point--to--point matching was close to the mean GSD value ($\approx 0.15$ m, the inverse of which was therefore used as a weighting for all correspondences within the Dynamic Network. 
   
    \item The inclusion of correspondences into the Dynamic Network effectively mitigated the ``wavy'' patterns in the point cloud registration caused by the attitude errors in trajectory. This was the largest source of georeferencing error for a pre-calibrated systems and nominal GNSS reception.
    
    \item The observered improvement in the point cloud registration was practically the same when employing only 5\% of the total number of correspondences; and still significant enhancement was detected using only 1\% to 0.1\% of correspondences. 
    
    \item Thanks to the correspondences, the point cloud registration via Dynamic Network was also able to compensate well with unknown boresight. As the accuracy of the recovered boresight parameters is related to their correlation with systematic errors in attitude, their determination was restrained by the geometry of the studied case of two parallel flight-lines. This had, however, very small impact on the residual errors in the point cloud due to the previously mentioned correlation.   
    
    \item Utilizing correspondences within the Dynamic Network can also effectively mitigate the imprecise point cloud registration during the absence of GNSS signals, especially when affecting only one of a pair of flight-lines.

\end{itemize}

In our opinion, methods that attempt to perform the integration of all the available information in a single step, such as the one put forward in this contribution, are to be preferred over multiple stages, cascade adjustment approaches because of their simplicity, rigour and effectiveness in correcting errors at the sensor measurement level, rather than in the intermediate products. The proposed approach can be applied to other kinematic scanning scenarios, such as terrestrial laser scanning, and generalise well to more complex multi-sensor adjustment problems.



\section*{Acknowledgment}
The helicopter flight and the reference sensors were provided by Helimap System. The personal involvement and contribution of its director Dr. Julien Valet are highly appreciated.
This contribution was partially supported by the Swiss National Foundation (200021-182072). The authors would like to thank Jesse LaHaye for his help in proofreading the manuscript.

\bibliography{References.bib}
\clearpage
\onecolumn
\section*{Appendix 1}
The following table summarizes for each tested descriptor in ALS its number of  dimension, parameters, run times and performances. 
Computation time is measured when estimating descriptors on a pair of fragments (of around $\sim$120'000 points each) and on a modern computer with the following configuration under Ubuntu 20.04:\\
Intel(R) Core(TM) i7-10750H CPU @ 2.60GHz, 32 Gb RAM, NVDIA RTX 2070.\\
The inlier ratio at 30 cm represents the fraction of retrieved correspondences whose error in the reference point clouds is below 30 cm. This is the main metric used to select the final descriptor.\\
For both FCGF and SpinNet, pre-trained version of their network on KITTI outdoor dataset are used. For LCD, the pre-trained version on indoor dataset is used since it is the only one available.
\begin{table}[ht]
\centering
\begin{tabular}{|c|c|c|c|c|c|} 
\hline
Descriptor & Dim. & Parameters                                                                                                         & \begin{tabular}[c]{@{}c@{}}Computation \\time [s]\end{tabular} & \begin{tabular}[c]{@{}c@{}}Match \\time [s]\end{tabular} & \begin{tabular}[c]{@{}c@{}}Inlier ratio \\@30cm [\% ]\end{tabular}  \\ 
\hline
SHOT       & 352  & $r_s$ = 15 * GSD                                                                                                   & 49                                                             & 73                                                       & 13,3                                                                \\ 
\hline
FPFH       & 32   & $r_s$ = 15 * GSD                                                                                                   & 16                                                             & 1,4                                                      & 5,7                                                                 \\ 
\hline
USC        & 1980 & $r_s$ = 15 * GSD                                                                                                   & 82                                                             & 612                                                      & 5,2                                                                 \\ 
\hline
FCGF       & 32   & \begin{tabular}[c]{@{}c@{}}vox. size = 0.2\\ kernel size = 7\end{tabular}                                          & 0,21                                                           & 1,3                                                      & 27,3                                                                \\ 
\hline
SpinNet    & 32   & \begin{tabular}[c]{@{}c@{}}$r_s$ = 2, rad$_n$ = 9, \\ele$_n$ = 30, azi$_n$ = 60\\vox$_r$ = 0.3, vox$_s$ = 30\end{tabular} & 1410                                                           & 2                                                        & 30,7                                                                \\ 
\hline
LCD        & 256  & \begin{tabular}[c]{@{}c@{}}$r_s$ = 2 m \\ patch pts N=1024\end{tabular}                                      & 128                                                            & 31                                                       & 53,7                                                                \\
\hline
\end{tabular}
\caption{\label{tab:desc_summary}Descriptors parameters, run-time and performances}
\end{table}
\end{document}